%% file: aaai25.tex
\documentclass[letterpaper]{article} 
\usepackage{aaai25}
\usepackage{times}  
\usepackage{helvet}  
\usepackage{courier}  
\usepackage[hyphens]{url}  
\usepackage{graphicx} 
\usepackage{natbib}  
\usepackage{caption} 
\usepackage{algorithm}
\usepackage{algorithmic}
\usepackage[utf8]{inputenc} 
\usepackage[T1]{fontenc}    
\usepackage{hyperref}       
\usepackage{url}            
\usepackage{booktabs}       
\usepackage{amsfonts}       
\usepackage{nicefrac}       
\usepackage{microtype}      
\usepackage{xcolor}         
\usepackage{multirow}
\usepackage{array}
\usepackage{amsmath}
\usepackage{xspace}
\usepackage{graphicx}
\usepackage{subcaption}
\usepackage{algorithm}
\usepackage{amssymb}
\usepackage{soul}
\usepackage[english]{babel}
\usepackage{todonotes}

\usepackage{newfloat}
\usepackage{listings}
\DeclareCaptionStyle{ruled}{labelfont=normalfont,labelsep=colon,strut=off} 
\floatstyle{ruled}
\newfloat{listing}{tb}{lst}{}
\floatname{listing}{Listing}
\title{Adaptive Draft-Verification for Efficient Large Language Model Decoding}

\author {
    Xukun Liu\textsuperscript{\rm 1},
    Bowen Lei\textsuperscript{\rm 2},
    Ruqi Zhang\textsuperscript{\rm 3}
    Dongkuan Xu\textsuperscript{\rm 4}
}
\affiliations {
    \textsuperscript{\rm 1}Northwestern University\\
    \textsuperscript{\rm 2}Texas A\&M University\\
    \textsuperscript{\rm 3}Purdue University\\
    \textsuperscript{\rm 4}North Carolina State University\\
    xukunliu2025@u.northwestern.edu, bowenlei@stat.tamu.edu, ruqiz@purdue.edu, dxu27@ncsu.edu
}

\usepackage{bibentry}

\begin{document}
\newcommand{\sysname}{\texttt{ADED}\xspace}
\definecolor{deepyellow}{rgb}{1.0, 0.8, 0.0}
\newcommand{\mcts}{\texttt{MCTS}\xspace}
\maketitle

\input{sections/1-abstract}

\input{sections/2-introduction}
\input{sections/3-method}

\input{sections/4-experiment}

\input{sections/5-conclusion}

\newpage

\bibliography{aaai25}
\newpage
\input{sections/6-appendix}

%
\begin{links}
    \link{Code}{https://anonymous.4open.science/r/ADED-C7D5/}
\end{links}


\end{document}

%% file: sections/1-abstract.tex
\vspace{-8mm}
\begin{abstract}

Large language model (LLM) decoding involves generating a sequence of tokens based on a given context, where each token is predicted one at a time using the model's learned probabilities. 
The typical autoregressive decoding method requires a separate forward pass through the model for each token generated, which is computationally inefficient and poses challenges for deploying LLMs in latency-sensitive scenarios.
The main limitations of current decoding methods stem from their inefficiencies and resource demands. Existing approaches either necessitate fine-tuning smaller models, which is resource-intensive, or relying on fixed retrieval schemes to construct drafts for the next tokens, which lack adaptability and fail to generalize across different models and contexts.
To address these issues, we introduce a novel methodology called \sysname \footnote{Project repo: \href{https://anonymous.4open.science/r/ADED-C7D5}{https://anonymous.4open.science/r/ADED-C7D5}}, which accelerates LLM decoding without requiring fine-tuning. Our approach involves an adaptive draft-verification process that evolves over time to improve efficiency. We utilize a tri-gram matrix-based LLM representation to dynamically approximate the output distribution of the LLM, allowing the model to adjust to changing token probabilities during the decoding process. Additionally, we implement a draft construction mechanism that effectively balances exploration and exploitation, ensuring that the drafts generated are both diverse and close to the true output distribution of the LLM.
The importance of this design lies in its ability to optimize the draft distribution adaptively, leading to faster and more accurate decoding. Through extensive experiments on various benchmark datasets and LLM architectures, we demonstrate that \sysname accelerates the decoding process while maintaining high accuracy, making it suitable for deployment in a wide range of practical applications.

\end{abstract}


%% file: sections/2-introduction.tex
\section{Introduction}


Large language model (LLM) decoding involves generating a sequence of tokens based on a given context, where each token is predicted one at a time using the model’s learned probabilities~\citep{llm1, llm2, llama, llama2}. The core mechanism is autoregressive, where each new token is generated conditioned on the previously generated tokens and the given context. This process is crucial for applications like text generation~\citep{text-generate,text-generate2,intro-gen}, machine translation~\citep{llm-mt, into-trans1, into-trans2}, and conversational AI~\citep{intro-con,intro-con1,intro-con2}. However, 
each decoding step involves a forward pass through the model, making the process inherently sequential and computationally expensive. The inefficiencies arise due to the need to reload the model for each token prediction, leading to high computational costs and memory bandwidth usage. 
This serial nature of decoding is a significant bottleneck, especially for real-time applications~\citep{real1,real2,real3} where latency is critical. Thus, optimizing the decoding speed of LLMs is essential for practical deployment.

\begin{figure}[t]
    \centering
    \includegraphics[width=0.5\textwidth]{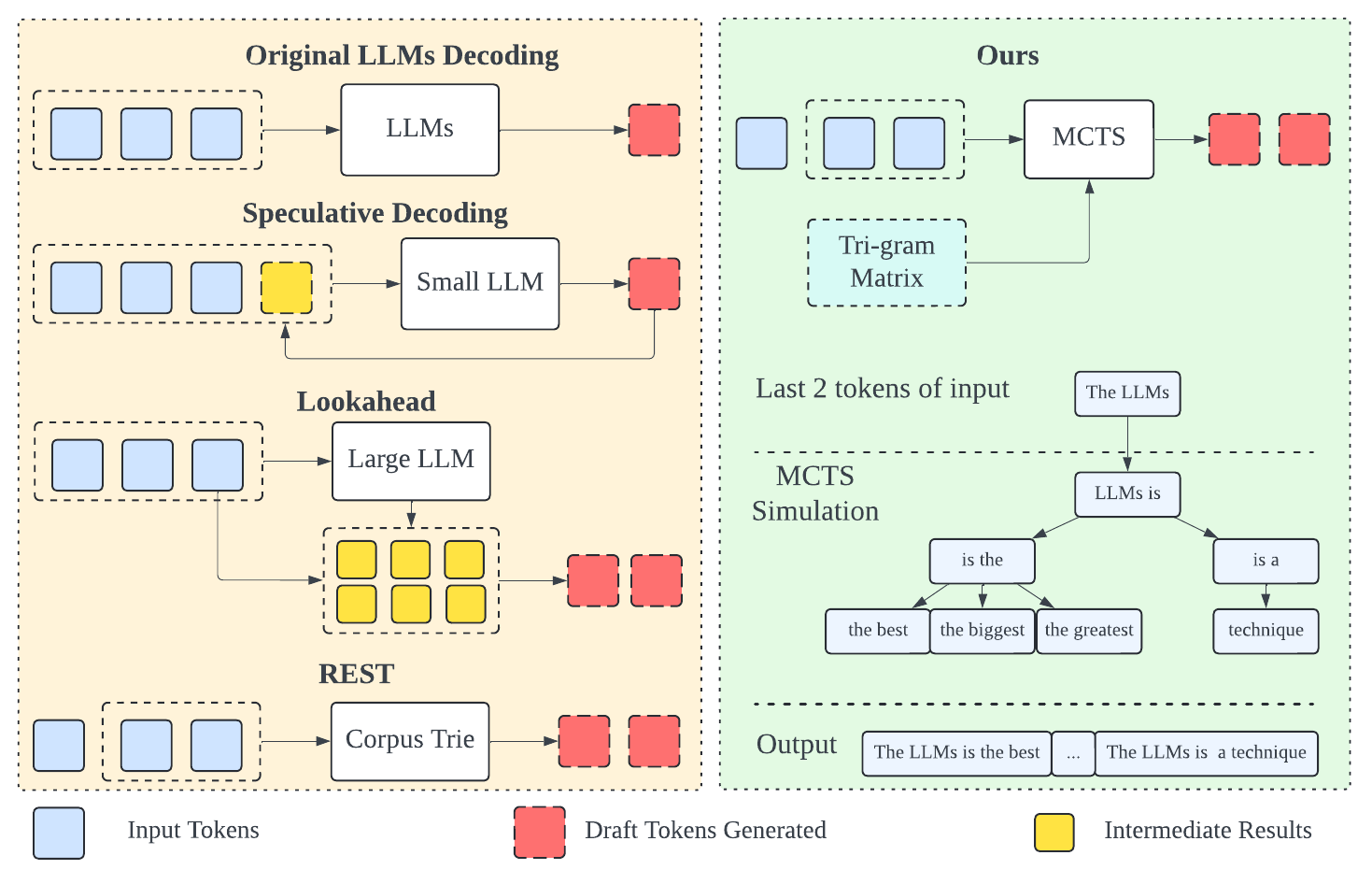}
    \vspace{-1mm}
    \caption{Comparison of different LLM decoding strategies. In \textit{Speculative Decoding}, a small LLM generates predictions (\textcolor{red}{red blocks}) from inputs (\textcolor{blue}{blue blocks}). \textcolor{deepyellow}{Yellow blocks} indicating intermediate results obtained from language model. \textit{Lookahead} uses a large LLM for forward-looking predictions. \textit{REST} employs a corpus trie for rapid token lookups. \sysname integrates Monte Carlo Tree Search with tri-gram statistics and recent token history to simulate potential outputs, refining its recommendations over time. \sysname's adaptive approach offers advantages in terms of speed and accuracy by continuously evolving its draft constructions.}
    \label{fig:overflow1}
    \vspace{-3mm}
\end{figure}

Recent research has explored various strategies to mitigate the inefficiencies of LLM decoding. \textit{Speculative Decoding}~\citep{specd,spec1,spec2} introduces an approach where a smaller, more efficient model generates several token predictions in parallel, which are then verified by the larger target model. This method leverages the efficiency of smaller models to reduce the number of serial forward passes required, achieving substantial speedups without altering the output distribution. \textit{Lookahead Decoding}~\cite{look} uses the full context to predict multiple future tokens, creating a buffer that reduces the dependency on sequential processing. \textit{REST}~\cite{he-etal-2024-rest} employs a retrieval-based approach where relevant tokens are fetched from a pre-constructed datastore using the current context, forming drafts that are verified by the LLM.
These methods can be summarized within the \textbf{\textit{draft-verification}} pipeline, as shown in Figure~\ref{fig:overflow1}. \textit{Speculative Decoding} and \textit{Lookahead} Decoding both generate draft tokens through predictive models, while REST constructs drafts from retrieved tokens based on the context. In each case, the drafts are then verified by the main LLM, ensuring that the final output adheres to the model's learned probabilities. Despite their advancements, these approaches face notable limitations. They often require additional training or fine-tuning, which can be resource-intensive. Fixed retrieval schemes lack adaptability, making it challenging to adjust the draft distribution in real-time based on the evolving LLM output. Additionally, these methods may not generalize well across different models and contexts, limiting their effectiveness in dynamic environments.




In this work, our focus is on \textbf{\textit{fine-tuning-free draft-verification}} to address these limitations. The draft-verification pipeline can be viewed as a rejection sampling procedure where the similarity between the proposal distribution (\textit{\ul{draft}}) and the target distribution (\textit{\ul{LLM output}}) is crucial for the acceptance rate and convergence speed. Higher similarity results in a higher acceptance rate and faster decoding speed. 
Very few fine-tuning-free approaches, \textit{e.g.}, \textit{REST}~\cite{he-etal-2024-rest}, typically use fixed retrieval-based schemes to construct drafts. These schemes lack the adaptability to adjust the draft distribution based on the evolving LLM output distribution, resulting in a persistent gap between the draft and the actual LLM output. This gap reduces the draft acceptance rate and limits the potential for improving decoding speed.
To address this issue, we raise the following question:





\textbf{\textit{Research Question:} \textit{How to design an adaptive draft construction process that can evolve itself and accurately approximate LLM outputs during decoding?} }

To introduce adaptability and find drafts that are increasingly close to the LLM output distribution during decoding, we not only need to have an adaptive draft construction pipeline but also need to maintain a balance between exploration and exploitation.
This balance ensures that speedups can be achieved by leveraging existing knowledge of draft construction while continuously exploring better draft construction capabilities.
To achieve this, we propose a novel methodology called \sysname (\textbf{\ul{A}}daptive \textbf{\ul{D}}raft-Verification for \textbf{\ul{E}}fficient LLM \textbf{\ul{D}}ecoding). \sysname incorporates a tri-gram-matrix-based adaptive LLM representative to control the conditional probability distribution of the next token, which can be updated during the decoding process to adjust the draft construction accordingly. 
To balance exploration and exploitation, we design a \textit{draft maker} inspired by \textit{Monte Carlo Tree Search (\mcts)}~\citep{mcts,mcts1,mcts2,mcts3}. This \textit{draft maker} uses a token preference score to maintain the balance during the search process. The score consists of two parts: the first part is based on the approximate conditional probability distribution of the next token obtained from the LLM representative, reflecting the \textit{draft maker}'s current knowledge of the LLM output; the second part encourages the \textit{draft maker} to explore unexplored or less-explored draft spaces.
Theoretically, we show that our method can be viewed as a constrained optimization problem to encourage the draft distribution to converge to the LLM output distribution.
Using the token preference score, the \textit{draft maker} can effectively search the draft space and generate candidate tokens. 
After the draft construction and verification are completed, the information is fed back to the LLM representative to update its approximation of the LLM output. This feedback loop enriches the \textit{draft maker}'s knowledge in subsequent rounds of draft-verification, enabling adaptability and self-improvement in the draft construction process.

In summary, our contributions are concluded as follows:

\begin{itemize}
    \item We design a \textit{tri-gram matrix-based} representation that dynamically approximates the LLM output distribution, enhancing adaptability without the need for fine-tuning. It addresses the limitation of fixed retrieval schemes by continuously evolving with the model’s predictions.
    \item We develop a \textit{draft maker} that effectively balances exploration and exploitation to generate high-quality drafts. This mechanism improves decoding speed and accuracy by ensuring that the drafts are closely aligned with the LLM’s output distribution. Our experiments show a \textbf{2.5X} improvement in decoding speed compared to baselines.
    \item Through extensive experiments on various benchmark datasets and LLM architectures, we demonstrate that \sysname successfully accelerates the decoding process while maintaining high accuracy. Specifically, we achieve up to a \textbf{2.5X} speedup in latency and an average acceptance rate improvement of \textbf{20\%} over existing methods.
    \item Our method’s ability to adapt to evolving LLM outputs and continuously refine draft construction sets it apart from existing ones, addressing the need for more flexible and dynamic decoding solutions.
\end{itemize}



%% file: sections/3-method.tex
\section{Methodology}
\begin{figure}[ht]
    \centering
    \includegraphics[width=0.49\textwidth]{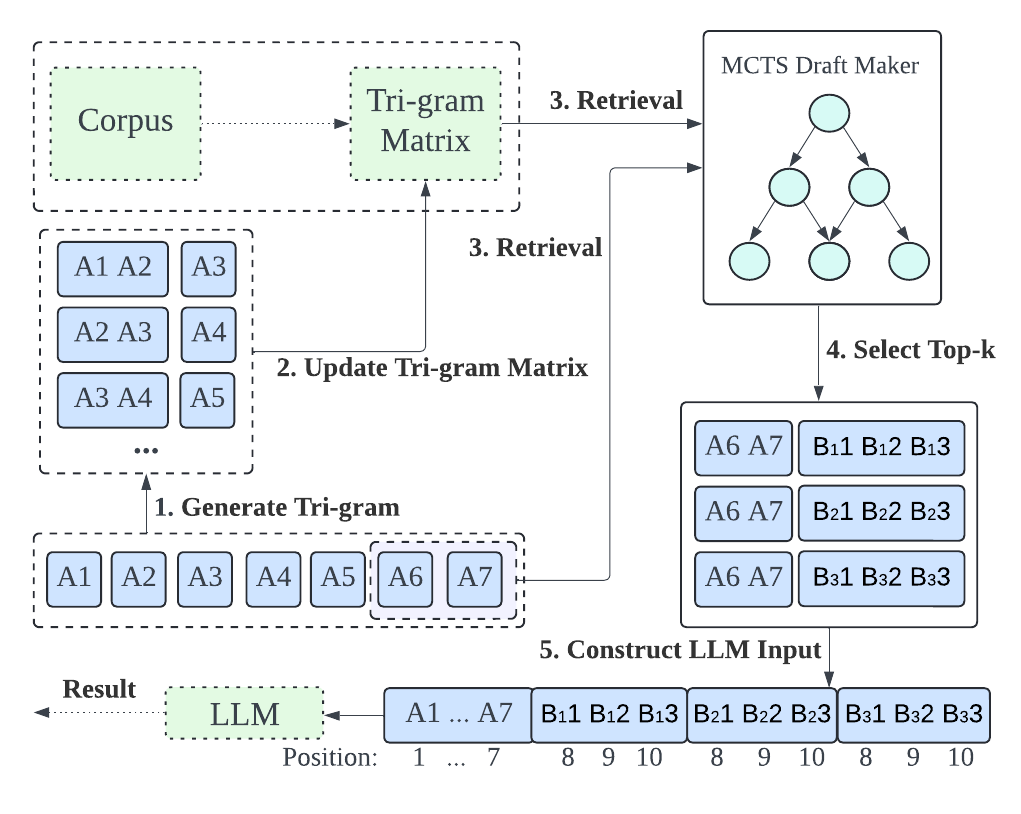}
    \vspace{-3mm}
    \caption{This figure illustrates the data processing workflow of \sysname. Initially, the input tokens undergo preprocessing to calculate their tri-grams, which serve to update the tri-gram matrix. Subsequently, the updated matrix, in conjunction with the last two tokens of the input, is used to retrieve potential token sequences. These sequences are ranked, and the top-k sequences are selected, and then appended to the original input. Finally, these extended sequences are inputted into the Large Language Model for prediction.}
    \label{fig:full_proc}
    \vspace{-3mm}
\end{figure}

We propose a new fast fine-tuning-free draft-verification LLM decoding method by introducing adaptability into the decoding and learning from LLM, which is illustrated in Figure~\ref{fig:full_proc}. Existing accelerated decoding algorithms either require additional fine-tuning or lack adaptability to LLM's output distributions, resulting in additional cost or insufficient acceleration. To address these issues, we design an adaptive LLM representation based on a tri-gram matrix to adaptively approximate the output distribution of the LLM, develop a draft maker that balances exploration and exploitation for self-improvement towards high-quality drafts, and verify the drafts using tree attention.

\subsection{Preliminary}
\textbf{\textit{Speculative decoding}} 
is a method to accelerate language model inference by using a smaller auxiliary model to generate a draft sequence, reducing the computational load on the larger model~\cite{specd}. \textbf{\textit{Retrieval-based speculative decoding}} extends this by incorporating a retrieval system instead of the smaller model, leveraging pre-stored corpus segments for relevant text generation. \textbf{\textit{Monte Carlo Tree Search}} (\mcts)~\citep{mcts,mcts1,mcts2,mcts3} is an algorithm that optimizes decision-making by balancing exploration and exploitation of future states. It selects nodes for further exploration using a combination of node visit counts and estimated values, aiming to maximize overall outcomes. For a comprehensive discussion of these methods, please refer to \textbf{\textit{Appendix E}}.

\subsection{Adaptive LLM Representative}
To approximate the output token distribution of the LLM without fine-tuning the small model, we distill linguistic knowledge from a small corpus and construct a tri-gram matrix as an initial representation of the LLM, which allows us to leverage the statistical regularities of language at a granular level. Specifically, we summarize and count each set of three tokens that appear in the corpus and compute the probability of the third token appearing conditional on the first two tokens. The formula is defined in Eq.~(\ref{tri-gram}):
\begin{equation}\label{tri-gram}
P(w_{i}|w_{i-2}, w_{i-1}) = \frac{C(w_{i-2}, w_{i-1}, w_{i})}{C(w_{i-2}, w_{i-1})},
\end{equation}
where $P(w_{i}|w_{i-2}, w_{i-1})$ is the conditional probability of a word $w_{i}$ given the two preceding words $w_{i-2}$ and $w_{i-1}$, $C(w_{i-2}, w_{i-1}, w_{i})$ is the count of the tri-gram occurrence in the corpus, and $C(w_{i-2}, w_{i-1})$ is the count of the preceding bi-gram~\cite{bi}.

In this way, we can obtain a good initial LLM representative at a much lower cost, which can generate an approximate distribution of the next token based on the previous tokens. This LLM representative will collaborate with our draft maker to generate drafts and get feedback to update the tri-gram matrix for adaptability and self-improvement. 
Please see \textbf{Section 2.3~\ref{sec-draft-maker}} for more details.

\subsection{Draft Maker and Self-Improvement}\label{sec-draft-maker}
With the help of the LLM representative, we further propose a draft maker that balances exploration and exploitation while searching for candidate drafts that are closer to the LLM output. On the one hand, the draft maker leverages the conditional probabilities from the LLM representative, which include current knowledge of the LLM output. On the other hand, the draft maker is encouraged to search more in the unexplored or less explored draft space to find better draft candidates.
Then, with the feedback from the LLM output, the LLM representative can update its understanding of the LLM output, improve the draft maker's search, and achieve self-improvement. Details are provided below.


\textbf{Draft Search Score}: 
Given the initial tokens, we exploit Monte Carlo Tree Search (\mcts)~\cite{mcts} to guide the search process of the drafts of the next tokens, where we prioritize candidate tokens according to the conditional probability from the tri-gram matrix-based LLM representative and the node visitation counts during the tree search. 
The score plays a key role in balancing exploration and utilization during the Monte Carlo tree search and is defined as Eq.~(\ref{eq:mcts}).
\begin{equation}\label{eq:mcts}
 \text{PUCT}(s, a) = Q(s,a) + E \cdot P(s,a) \cdot \frac{\sqrt{\sum_{b}N(s,b)}}{1 + N(s,a)}.
\end{equation}
The score design is motivated by PUCT Score~\citep{putc,alpha}. In particular, \(Q(s, a)\) assesses the quality of taking action \(a\) in state \(s\), while \(P(s, a)\) represents the prior probability of selecting action \(a\) in state \(s\). The term \(N(s, a)\) denotes the number of times the action \(a\) has been taken from state \(s\), and $\sum_{b}N(s,b)$ sums the counts for all actions from state \(s\). 
Eq.~(\ref{eq:e}) plays a critical role in determining the balance between exploration and exploitation within the \mcts framework. 
\begin{equation}\label{eq:e}
E = C_1+\log\left(\frac{\sum_{b}N(s,b) + C_2 + 1}{C_2}\right),
\end{equation}
The constant $C_1$ acts as a base level adjustment, while $C_2$ modulates the logarithmic term to scale the exploration factor dynamically based on the total visitation counts. This formula ensures that our draft choices are contextually appropriate and optimizes the robustness and coherence of text generation.





\textbf{Self-Improvement Strategy Transfer}: Based on the final search score obtained during the 
search, we can construct draft candidates and verify them to get the final decoding output (please see \textbf{Section 2.4} \ref{subsec:draft-c-v}) and feed it back for self-improvement. This final output decoding represents LLM's output distribution, which would be a good learning material for the LLM representative. Therefore, we feed this knowledge into the LLM representative in order to obtain updated conditional probability distributions, thus providing the draft maker with more accurate and exploitable knowledge, which is illustrated in Figure~\ref{fig:full_proc}.
Specifically, this technique operates by first extracting tri-grams from recent outputs of the LLM. Each tri-gram's frequency is then used to update its probability as potential outputs. These adjusted probabilities are fed into the \mcts as part of the policy network, influencing the selection phase of the tree search. 
The updated tri-gram probabilities essentially serve as a dynamic policy guide, enhancing the model's ability to generate contextually relevant and coherent sequences. By incorporating learned tri-gram probabilities into the tree search algorithm, we effectively create a feedback loop where the search strategy itself evolves over time. This strategy adjustment is executed by recalibrating the exploration-exploitation balance based on the empirical data derived from the model's own outputs.

\subsection{Draft Construction and Verification}\label{subsec:draft-c-v}
It is important to note that candidate drafts generated by the draft maker often have common starting segments that can cause redundant recalculations in the Transformer layers if not managed correctly. 
To address the issue,
a pseudo-sequence that guarantees that each draft is a sub-sequence and that any common prefix appears only once is created~\cite{he-etal-2024-rest}. 
Motivated by this observation, we use a specific attention mask for each attention layer, 
called \textit{tree attention}~\citep{ta,medusa}. This mask aligns the computations for each token with its dependencies according to the original draft sequence, preserving the draft's contextual integrity and preventing unnecessary computations. The approval of drafts relies on a comparison with the conditional distribution from the LLM. At each position, new tokens are sampled and compared to the draft tokens. If a sampled token corresponds to the draft token, it is approved; otherwise, the draft is discarded from that point. This selective approval ensures that the output sequence aligns with what would be produced by a typical autoregressive process, thus upholding the authenticity of the generated text.


\section{Theoretical Insight: Why \sysname uses \mcts}

In this section, we provide a theoretical justification for the design of the \sysname method. We show that the draft search in \sysname using \mcts can be viewed as a form of policy optimization, while the inference mechanism of LLMs can be viewed as a similar form of penalty optimization.


\textbf{\mcts in \sysname}: The token selection procedure in \sysname decoding can be viewed as an action selection process. The \mcts algorithm optimizes its policy by iteratively building a search tree and updating visit counts for each node (state-action pair) based on the search paths. The visit count distribution $\hat{\pi}(a \mid x)$ is defined as:
\begin{equation}
\hat{\pi}(a \mid x) \triangleq \frac{1 + n(x, a)}{|A| + \sum_{b} n(x, b)},
\end{equation}
where $n(x, a)$ represents the visit count for action $a$ in state $x$, and $|A|$ represents the total number of possible actions at state $x$.  Then, the action selection in \mcts can be written as selecting the action $a^*$:
\begin{equation}
    a^*(x) \triangleq \arg\max_{a} [Q(x,a) + \lambda_N \cdot \frac{\pi_{\theta}(a \mid x)}{\hat{\pi}(a \mid x)}]
\end{equation}
Following~\citep{mctsRPO}, we use $q \in \mathcal{R}^{|A|}$ to denote the vector of Q-function $Q(x,a)$.
With proper choice of hyper-parameters, the \mcts algorithm can be viewed as searching for the optimum solution to a policy optimization problem~\citep{mctsRPO} as below:
\begin{equation}
\bar{\pi} \triangleq \arg\max_{y \in S} \left[ q^\top y - \lambda_N \text{KL}[\pi_{\theta}, y] \right],\label{eq-mcts-op}
\end{equation}
where $S$ is the $|A|$-dimensional simplex, $\lambda_N$ is a regularization parameter that depends on hyperparameters and balances exploration and exploitation, and $\text{KL}$ is the KL-divergence.

\textbf{LLM Inference Mechanism}: Large language models, particularly those based on the Transformer architecture, generate text by predicting the probability distribution of the next token given the previous tokens. During training, the model maximizes the log-likelihood of the observed data, which is equivalent to minimizing the cross-entropy loss:
\begin{equation}
\mathcal{L}(\theta) = -\sum_{t=1}^{T} \log P(w_t \mid w_{1:t-1}; \theta) + \frac{\lambda}{2} \|\theta\|_2^2,
\label{eq-llm-infe}
\end{equation}
where $P$ denotes the conditional probability of LLM, $w$ denotes the tokens, and $\theta$ denotes the model parameters.

\begin{table*}[t]
  \centering
  \vspace{-2mm}
  \caption{Latency and Average Accept Length Comparison between \sysname and Baselines. In most test cases, \sysname has the lowest latency, longer accept length, and higher efficiency.}
  \resizebox{0.99\textwidth}{!}{
  \begin{tabular}{ccccccccccc}
    \toprule &
    & \multicolumn{5}{c}{Latency} & \multicolumn{4}{c}{Average Accept Length} \\
    \cmidrule(lr){3-7} \cmidrule(lr){8-11}
    Benchmark & Model & REST & REST Single & Lookahead & Autoregressive & \sysname & REST & REST Single & Lookahead & \sysname\\
    \midrule 
    \multirow{5}{*}{MT-Bench} &
    Vicuna-7B & 16.31 & 17.36 & 18.93 & 24.77 & \textbf{12.95}  & 1.97 & 1.98 & 1.89 &  \textbf{2.42} \\
    & Vicuna-13B & 25.43 & 25.99 & 32.73 & 44.07 & \textbf{22.94} & 1.98 & 1.99 & 1.85 & \textbf{2.39} \\
    & Vicuna-33B & 28.63 & 28.62 & 40.53 & 52.97 & \textbf{24.96} & 1.95 & 1.96 & 1.83 &  \textbf{2.29}\\
    & Llama2-7B & 16.08 & 17.67 & 18.84 & 25.58 & \textbf{13.85} & 1.96 & 1.95 & 1.96  &  \textbf{2.30}\\
    & Llama2-13B & 27.13 & 29.80 & 31.24 & 44.76 & \textbf{25.13} & 1.95 & 1.95 & 1.96 &  \textbf{2.32} \\
    \midrule 
    \multirow{5}{*}{Alpaca} &
    Vicuna-7B & 14.24 & 14.58 & 18.73 & 24.49 & \textbf{12.81} & 2.22 & 2.22 & 1.89 &  \textbf{2.33} \\
    & Vicuna-13B & \textbf{22.94} & 23.01 & 32.60 & 43.60 & 24.06 & 2.21 & 2.21 & 1.86 & \textbf{2.26} \\
    & Vicuna-33B & 26.03 & 25.89 & 40.58 & 52.52 & \textbf{24.62} & 2.11 & 2.12 & 1.82 &  \textbf{2.21}\\
    & Llama2-7B & 14.13 & 14.87 & 19.28 & 25.38 & \textbf{12.90} & 2.21 & 2.20 & 1.97 &  \textbf{2.37}\\
    & Llama2-13B & 23.66 & 24.07 & 31.18 & 44.04 & \textbf{23.57} & 2.15 & 2.13 & 1.96 &  \textbf{2.32} \\
    \midrule
    \multirow{5}{*}{Human Eval} &
    Vicuna-7B & 14.90 & 15.56 & 18.99 & 25.49 & \textbf{11.24} & 2.21 & 2.23 & 2.10 &  \textbf{2.67} \\
    & Vicuna-13B & 20.17 & 20.61 & 27.43 & 45.13 & \textbf{19.96} & 2.50 & 2.50 & 2.23 & \textbf{2.81} \\
    & Vicuna-33B & 24.91 & 25.06 & 31.34 & 52.32 & \textbf{21.19} & 2.29 & 2.30 & 2.02 &  \textbf{2.62}\\
    & Llama2-7B & 14.37 & 15.57 & 15.28 & 25.91 & \textbf{11.68} & 2.19 & 2.19 & 2.27 &  \textbf{2.63}\\
    & Llama2-13B & 25.46 & 25.85 & 26.72 & 45.25 & \textbf{21.82} & 2.01 & 2.01 & 2.17 &  \textbf{2.60} \\
    \bottomrule
  \end{tabular}
  }
  \label{main_lat}
  
\end{table*}
\begin{figure*}[!t]
\vspace{-0mm}
    \begin{subfigure}[b]{0.69\columnwidth}
        \includegraphics[width=\textwidth]{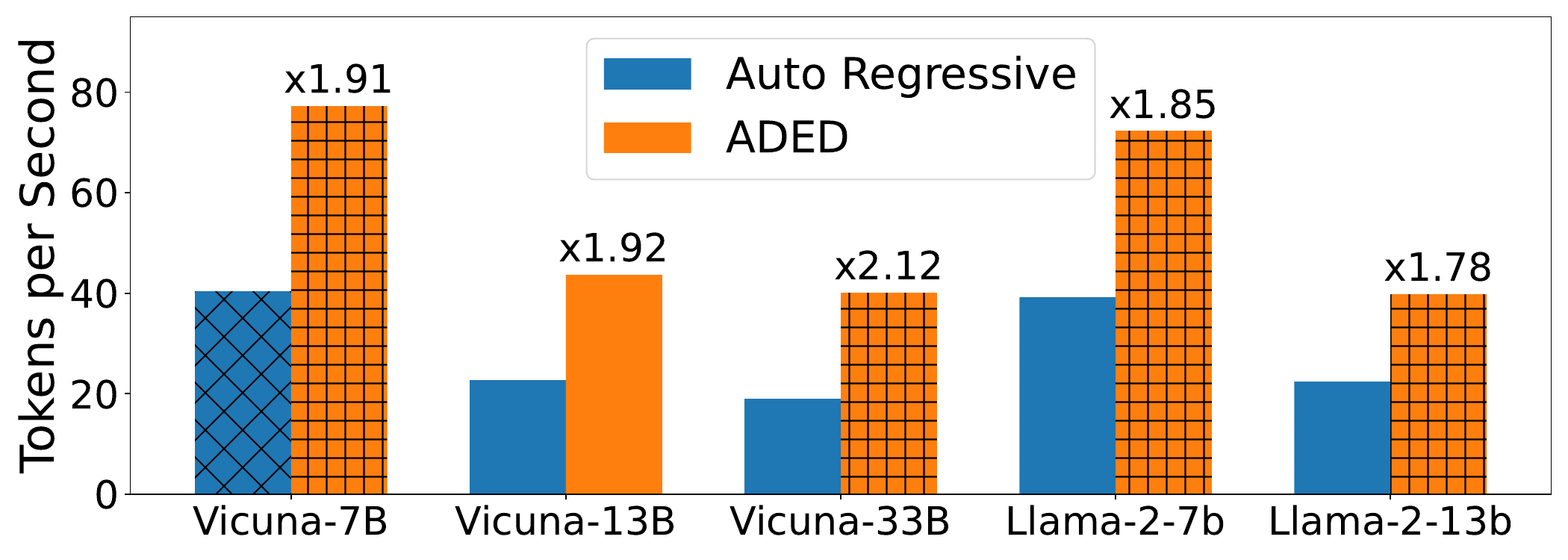}
        \caption{Throughput and Speedup on MT-Bench}
        \label{fig:sub1}
    \end{subfigure}
    \begin{subfigure}[b]{0.69\columnwidth}
        \includegraphics[width=\textwidth]{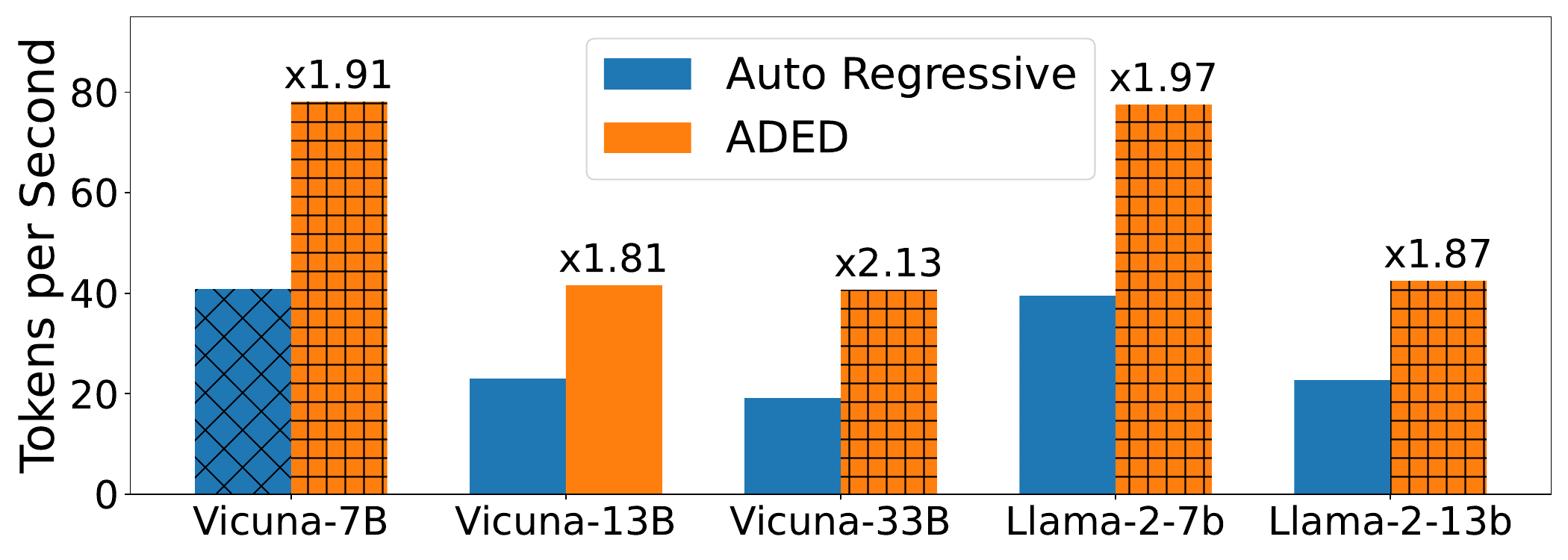}
        \caption{Throughput and Speedup on Alpaca}
        \label{fig:sub2}
    \end{subfigure}
    \begin{subfigure}[b]{0.69\columnwidth}
        \includegraphics[width=\textwidth]{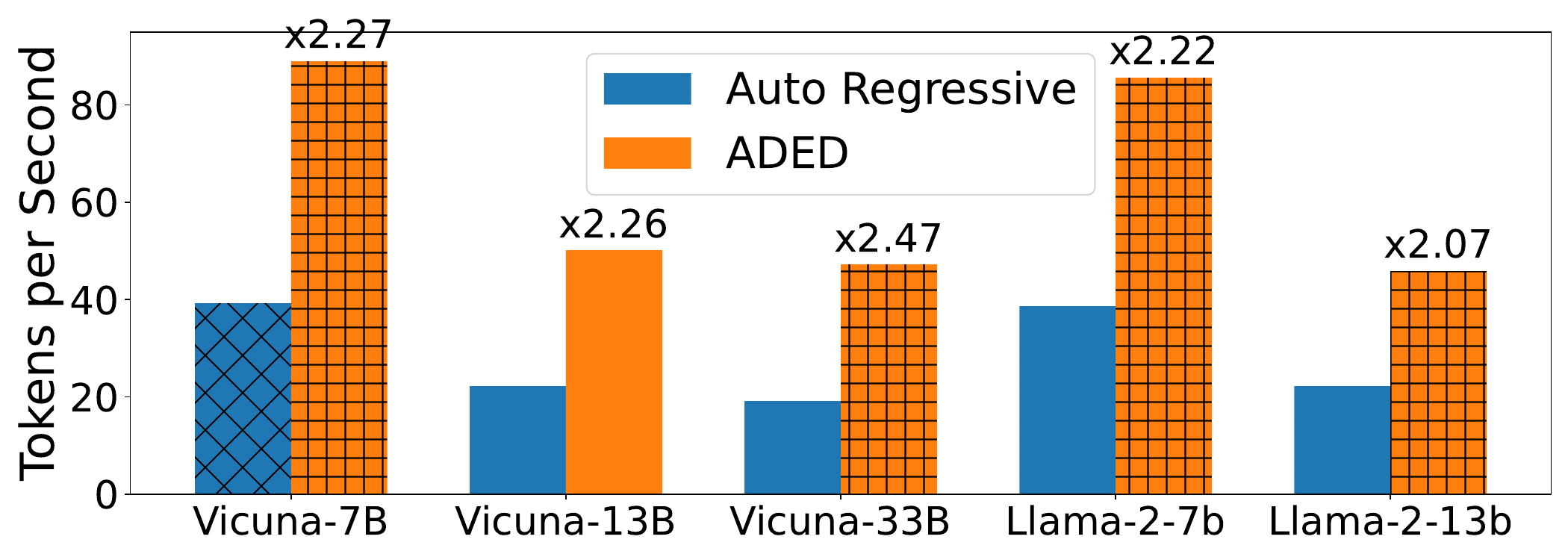}
        \caption{Throughput and Speedup on Human-Eval}
        \label{fig:sub3}
    \end{subfigure}
    \caption{Comparison of \sysname's throughput for different models on (a) MT-Bench, (b) Alpaca, and (c) Human-Eval. The performance of \sysname shows stable and significant improvements across different models and benchmarks.}
    \label{fig:all_speed_123}
    \vspace{-3mm}
\end{figure*}

\textbf{Comparative Analysis}:
As shown in Eq~(\ref{eq-mcts-op}) and Eq.~(\ref{eq-llm-infe}), both \mcts and LLMs can be viewed as regularized optimization problems for selecting the distribution of the next tokens. On the one hand, the Q-function in \mcts for \sysname can be viewed as an approximation to the log-likelihood of LLMs:
%
\begin{align}
    Q(x,a) &= -\sum_{t=2}^{T} \log \hat{P}(w_t \mid w_{t-1},w_{t-2}; \theta) \nonumber \\
    &  \approx \log P(w_0, w_1, \cdots, w_T; \theta) \nonumber \\
    & = -\sum_{t=2}^{T} \log P(w_t \mid w_{1:t-1}; \theta),
\end{align}
where $\hat{P}$ and $P$ are the conditional probability distribution from tri-gram-matrix-based LLM representative and  LLMs, respectively. On the other hand, both \mcts and LLMs employ regularization
to improve the optimization procedure.
As a result, we verify the similarities between \mcts and LLM Inference in terms of optimization and regularization.

%% file: sections/4-experiment.tex
\section{Experiments}

\subsection{Experimental Setup}

\textbf{Models and Datasets.} 
We conduct a series of experiments with five distinct models on three datasets to evaluate the efficacy of \sysname.
In particular, We 
use three Vicuna models~\cite{vicuna} (7B, 13B, 33B) and two LLaMA2-chat models~\cite{llama2} (7B, 13B) to evaluate the acceleration capabilities across different model sizes and types. Our assessment incorporates the HumanEval~\cite{human-eval}, MT-Bench~\cite{mtbench}, and Alpaca~\cite{alpaca} datasets to ascertain general natural language understanding and generation competencies. These datasets are meticulously chosen to guarantee a comprehensive analysis of the acceleration techniques across various tasks.

\textbf{Corpus.} We construct two corpora. The first one is built using a portion of the Python pre-training code from The Stack~\cite{stack}, comprising about 2.7M Python code samples with a resulting size of 1007MB. The second is constructed using data derived from UltraChat~\cite{ultrachat}, consisting of around 774K ChatGPT conversations, producing a corpus with a size of 574MB. The experiments on the MT-Bench and Alpaca are conducted using the UltraChat corpus, while the Human-Eval benchmark utilize the corpus from The Stack.

\textbf{Metrics.} To assess the acceleration performance
on large language models, we use two main metrics: speedup ratio and average acceptance length. Speedup ratio, calculated as the ratio of the time required by the baseline models to complete inference tasks without acceleration to the time required by our \sysname, measures the efficiency gains introduced by the algorithm. The second metric, average acceptance length, measures the average number of tokens accepted per forward pass by the target large language models, excluding any overhead of retrieving and constructing draft tokens, indicating the maximum possible acceleration.

\textbf{Baselines.} 
We compare various foundational approaches to improve the decoding speed of large language models. We examine \textit{Lookahead Decoding}~\cite{look}, a precise and parallel decoding algorithm that cuts down latency without relying on draft models. We compare \textit{REST}~\cite{he-etal-2024-rest} (Retrieval-Based Speculative Decoding), which adopts a retrieval-based strategy to create draft tokens, in contrast to conventional speculative decoding methods that rely on a draft model.For fairness in comparison, we include \textit{REST Single}, a single-threaded version of REST, to evaluate performance under constrained processing conditions. We also include the traditional \textit{Autoregressive} method, which represents the standard decoding approach, serving as a baseline to highlight the improvements offered by the other methods.
All experiments are conducted on NVIDIA A6000 GPUs, except for the 33B model, which utilizes an NVIDIA H100.
The experiments default to Greedy sampling.

\textbf{Configuration of \sysname.}
 To ensure reproducibility, we provide detailed hyperparameters, the corpus used, and the runtime environment for each set of experiments in \textbf{\textit{Appendix F}}. This allows interested researchers to replicate our findings accurately and compare them against their setups. Due to space constraints, these details are not included in the main text but are comprehensively documented in the appendix.

\subsection{Main Results}
In the experiments, we compare the efficacy of different baselines applied to various models, utilizing three datasets: MT-Bench, Human-Eval, and Alpaca. We focus on metrics of Accept Length, Latency, and Speedup Ratio.
Table~\ref{main_lat} summarizes the latency and average accept length  
on the three datasets. \sysname consistently demonstrates lower latency, 
particularly for the vicuna-7B and llama2-13B models. For instance, on MT-Bench, \sysname achieves a latency of 12.95 ms for vicuna-7B, which is lower than \textit{REST} (16.31 ms), \textit{REST} Single Thread (17.36 ms), and \textit{Lookahead} (18.93 ms). 
Notably, \textbf{the memory required for \sysname (574MB) is only 5.6\% of that required for REST (12GB). According to Table \ref{tab:corpus_size}, even when \sysname uses a smaller corpus (253MB, which is just 2.5\% of REST's requirements), it still achieves lower latency than REST}. This trend is also observed on Alpaca, where \sysname achieves a latency of 12.81 ms for vicuna-7B, compared to 14.24 ms for REST, 14.58 ms for \textit{REST} Single Thread, and 18.73 ms for \textit{Lookahead}.

The accept length results in Table~\ref{main_lat} indicate the quality of the generated outputs, with longer accept lengths suggesting more coherent and contextually relevant text. Our method, \sysname, outperforms other methods across different models on both MT-Bench and Alpaca datasets. For example, on MT-Bench, \sysname achieves the highest accept length for vicuna-33B and llama2-13B models, showcasing its superior language generation capabilities.

Speedup ratio are used to evaluate the efficiency. 
\sysname consistently shows a significant improvement in speed up across all datasets in Figure ~\ref{fig:all_speed_123}.
This efficiency is noticeable on the MT-Bench, Alpaca and Human-Eval datasets, where \sysname not only reduces latency but also enhances the overall processing speed.
For instance, \sysname achieves a speedup of 1.92x on MT-Bench with the vicuna-13B model, outperforming \textit{REST}, \textit{REST} Single Thread, and \textit{Lookahead}. On the HumanEval dataset, the vicuna-33B model, for example, demonstrates a speedup of nearly 2.5x when using \sysname.



\subsection{Stability of \sysname}
\begin{figure}[t]
\vspace{-3mm}
    \centering
    \begin{subfigure}[b]{0.55\columnwidth}
        \includegraphics[width=\textwidth]{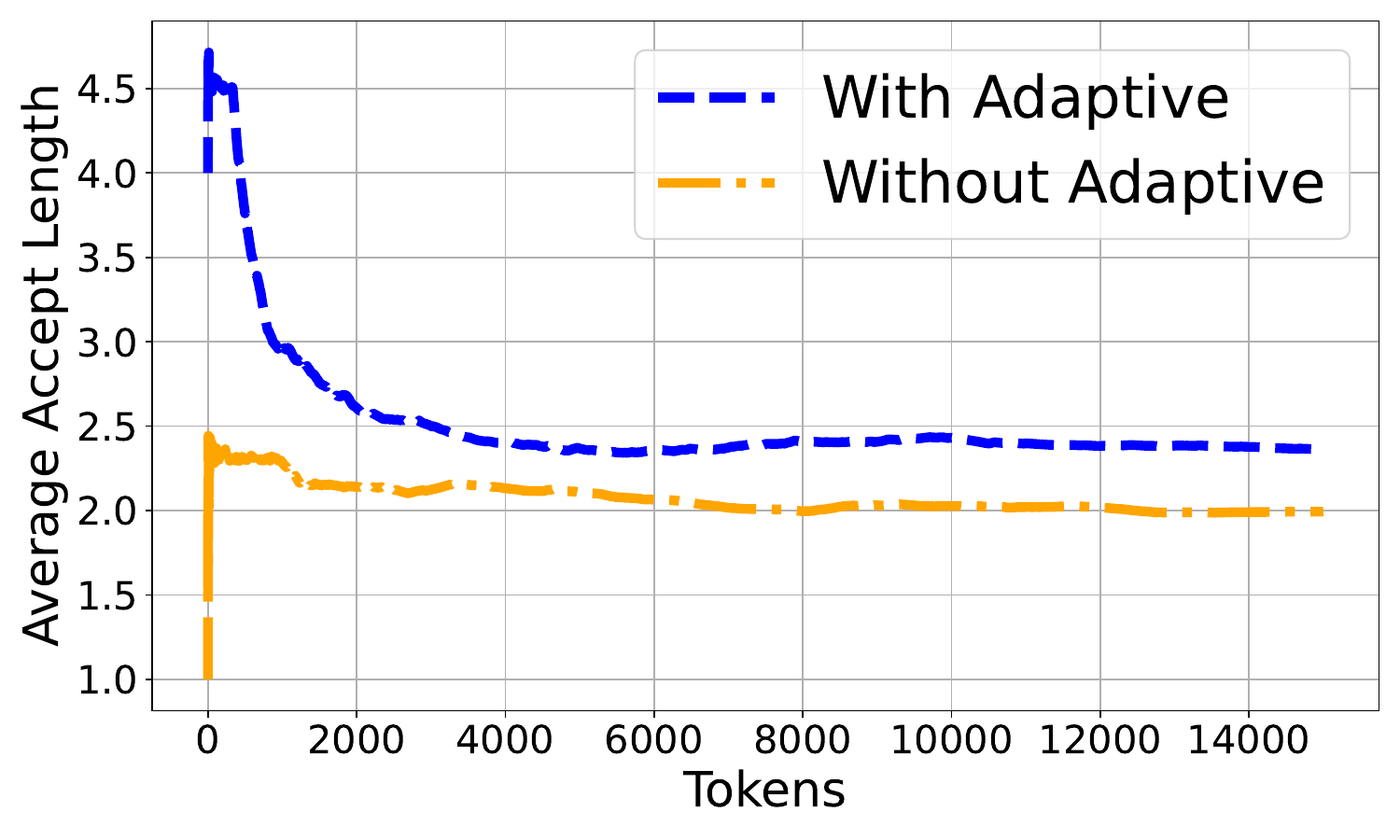}
        \vspace{-3mm}
        \caption{Effect of the adaptive strategy}
        \label{fig:adaptive}
    \end{subfigure}
    \begin{subfigure}[b]{0.4\columnwidth}
        \includegraphics[width=\textwidth]{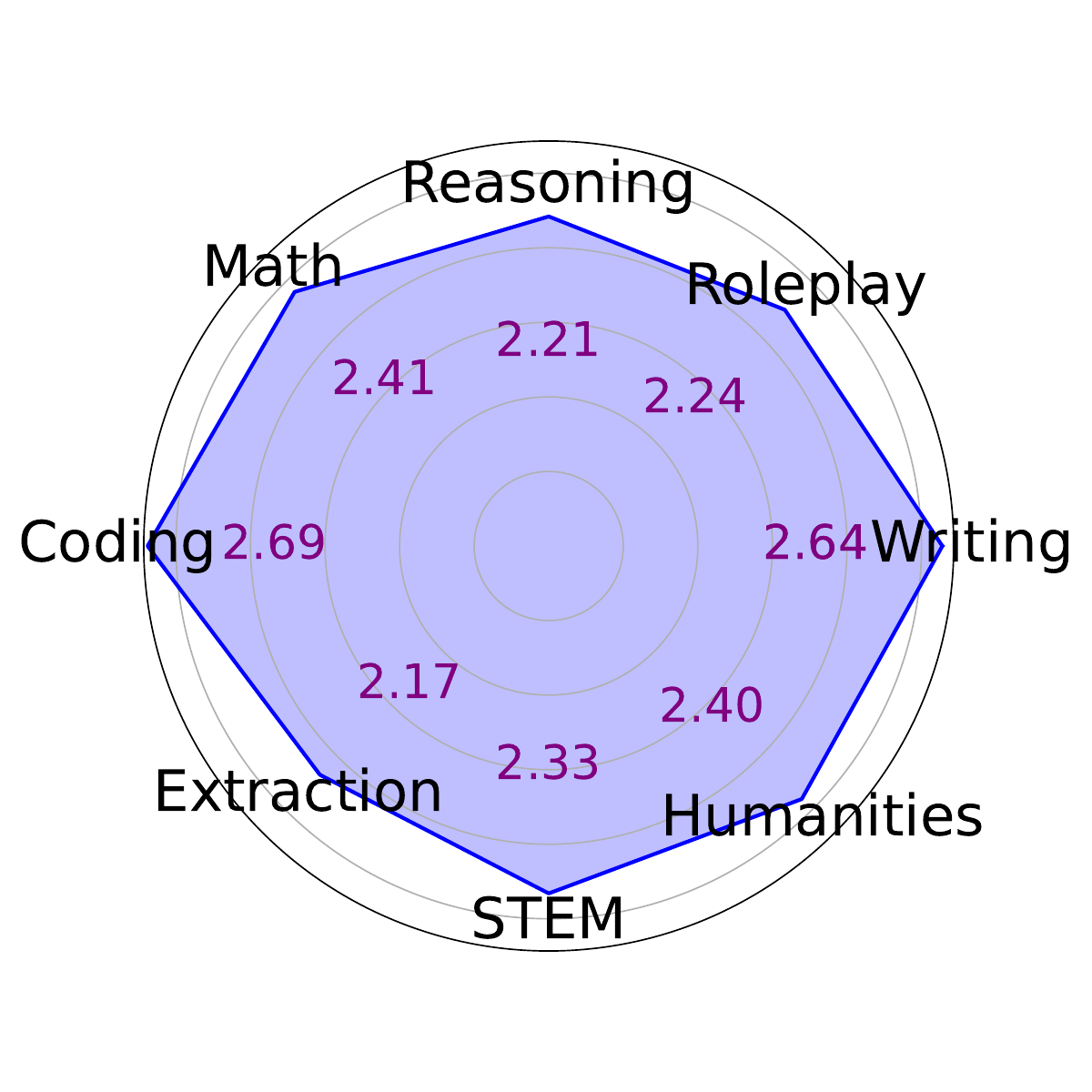}
        \vspace{-3mm}
        \caption{Stability amount different tasks}
        \label{fig:task_stability}
    \end{subfigure}
    \vspace{-0mm}
    \caption{(\ref{fig:adaptive})~Adaptive Strategy comparison on MTBench: Performance of Vicuna-7B model with and without the adaptive strategy on the MT-Bench dataset, showing the advantage of using the adaptive approach. (\ref{fig:task_stability})~Average Accept Length for different tasks on MT-Bench, demonstrating that \sysname consistently performs well across tasks.}
    \label{fig:both_}
    \vspace{-3mm}
    
\end{figure}
In this section, we analyze the stability of 
\sysname across different categories of tasks. The categories include writing, roleplay, reasoning, math, coding, extraction, STEM, and humanities. The experimental results shown in Figure~\ref{fig:task_stability} indicate that \sysname maintains consistent performance across categories. The average accept length remains stable, demonstrating that \sysname can effectively handle a diverse range of tasks without significant variations in performance.

\begin{figure}[t]
\centering
    \begin{subfigure}[b]{0.49\columnwidth}
        \includegraphics[width=\textwidth]{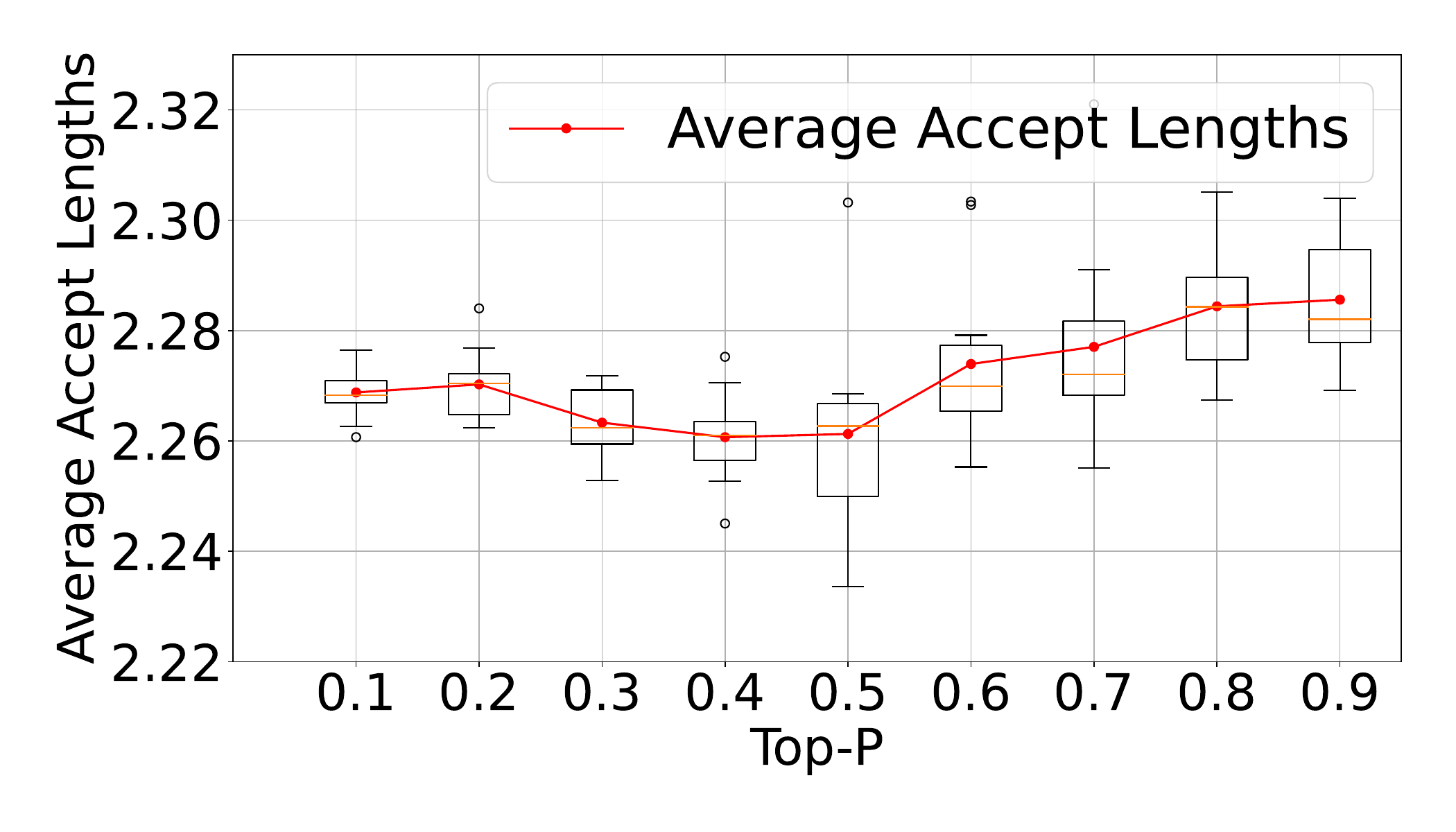}
        \vspace{-3mm}
        \caption{Top-p}
        \label{fig:top_p_sensitivity}
    \end{subfigure}
    \begin{subfigure}[b]{0.49\columnwidth}
        \includegraphics[width=\textwidth]{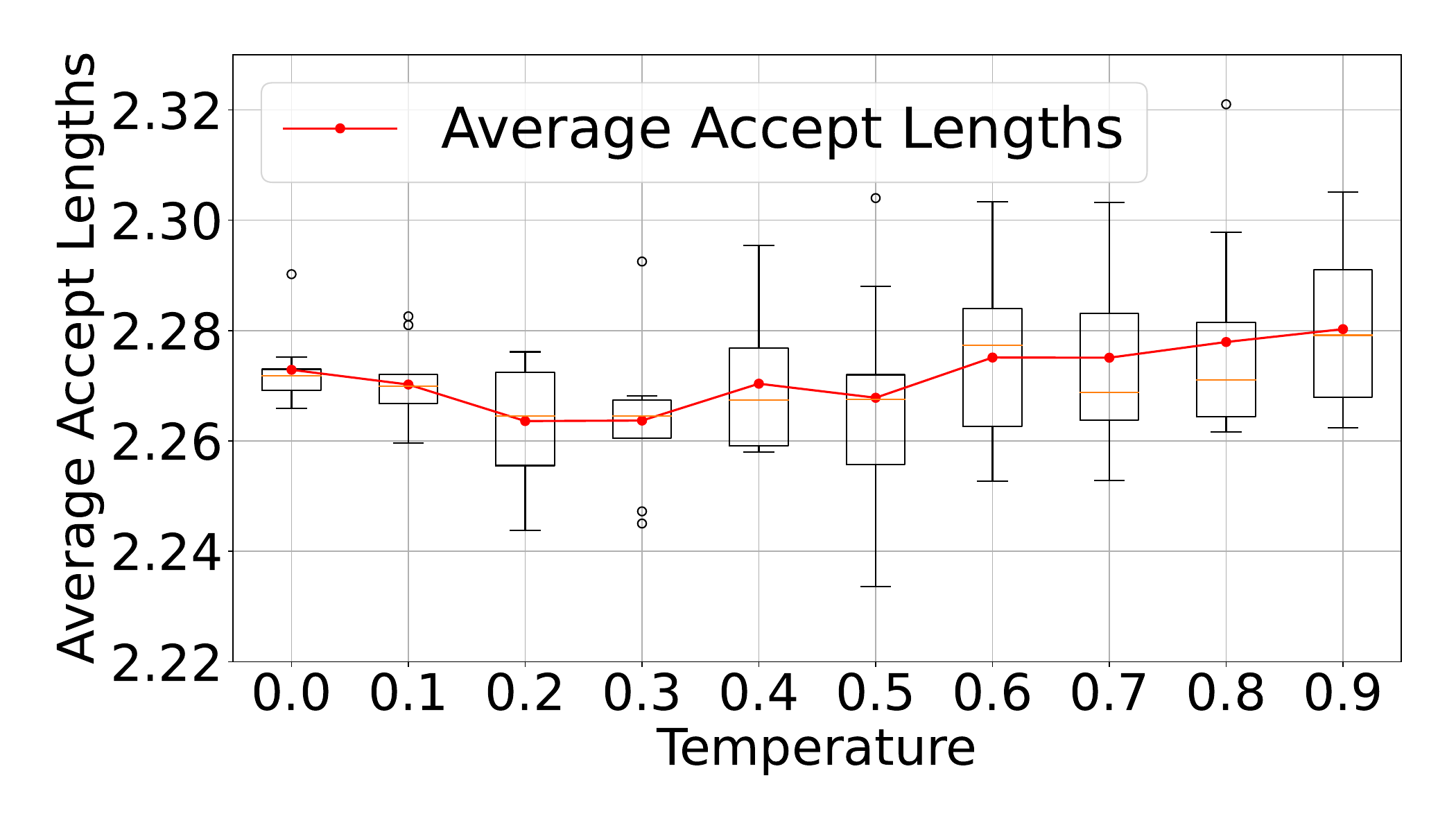}
        \vspace{-3mm}
        \caption{Temperature}
        \label{fig:temperature_sensitivity}
    \end{subfigure}
    \vspace{-0mm}
    \caption{Sensitivity analysis of \sysname on (\ref{fig:top_p_sensitivity}) top-p and (\ref{fig:temperature_sensitivity}) temperature parameters.}
    \label{fig:both}
    \vspace{-2mm}
\end{figure}

To further evaluate the robustness of \sysname, we examine the effects of varying the top-p and temperature parameters on the performance. Figures~\ref{fig:top_p_sensitivity} and~\ref{fig:temperature_sensitivity} summarize the impact of these parameters on the average accept length.
In particular, Figure~\ref{fig:top_p_sensitivity} shows that changes of top-p do not affect the performance of \sysname significantly. The average accept length remains relatively stable across different values of top-p, indicating that \sysname is not overly sensitive to this parameter.
Figure~\ref{fig:temperature_sensitivity} demonstrates that variations in the temperature parameter have negligible impact on the performance of \sysname. The consistency in the average accept length across different temperature values further supports the robustness.



    

\section{Ablation Study}

To gain insight into our method, we conduct a series of ablation studies. Full studies are summarized in \textbf{\textit{Appendix D}}.

\textbf{Effect of the adaptive strategy.}
Figure~\ref{fig:adaptive} illustrates the performance of our adaptive strategy on Vicuna-7B, with 
the analysis of average accept lengths over varying token counts. 
We find that the adaptive strategy maintains a higher average accept length over the entire range compared to the non-adaptive strategy.
The adaptive strategy's success is attributed to its dynamic adjustment of the model's probability distributions based on the tri-gram frequencies from prior outputs. This allows the model to better manage longer contexts and maintain relevance, enhancing stability and coherence in longer interactions. 

\textbf{Effect of the corpus size.}
Table~\ref{tab:corpus_size} 
shows the impact of the increase in corpus size from 121k to 774k on various performance metrics.
With the expansion of the corpus, there is a gradual improvement in the Accept Length from 2.30 to 2.42. This increase suggests that larger datasets provide a broader array of language patterns, which enhances the model's ability to generate more coherent and contextually relevant outputs. Despite the increase in data size from 253 MB to 574 MB, the system maintains efficient data processing capability. The small differences in latency affirm \sysname's consistent performance, even with smaller corpus sizes, which further extends its potential for use on resource-constrained devices.
\begin{table}[ht]
  \centering
  \caption{Effect of Corpus Size.}
  \vspace{-2mm}
  \resizebox{0.5\textwidth}{!}{
  \begin{tabular}{ccccc}
    \toprule
    Corpus Size & Corpus Size & Latency & Accept Length & Speed up \\
    \midrule
    121k & 253 MB & 13.09 ms & 2.30 & 1.89 \\
    \midrule
    774k & 574 MB & 12.95 ms & 2.42 & 1.93 \\
    \bottomrule
  \end{tabular}
  }
  \label{tab:corpus_size}
  \vspace{-3mm}
\end{table}
The modest rise in retrieval time underscores the efficiency of the retrieval algorithms, which can manage larger datasets without significantly compromising response speed. In summary, the results show that larger corpus sizes can improve the quality of the model's output while maintaining good system performance.

\begin{figure}[htb]
    \centering
    \begin{subfigure}[b]{0.49\columnwidth}
        \includegraphics[width=\textwidth]{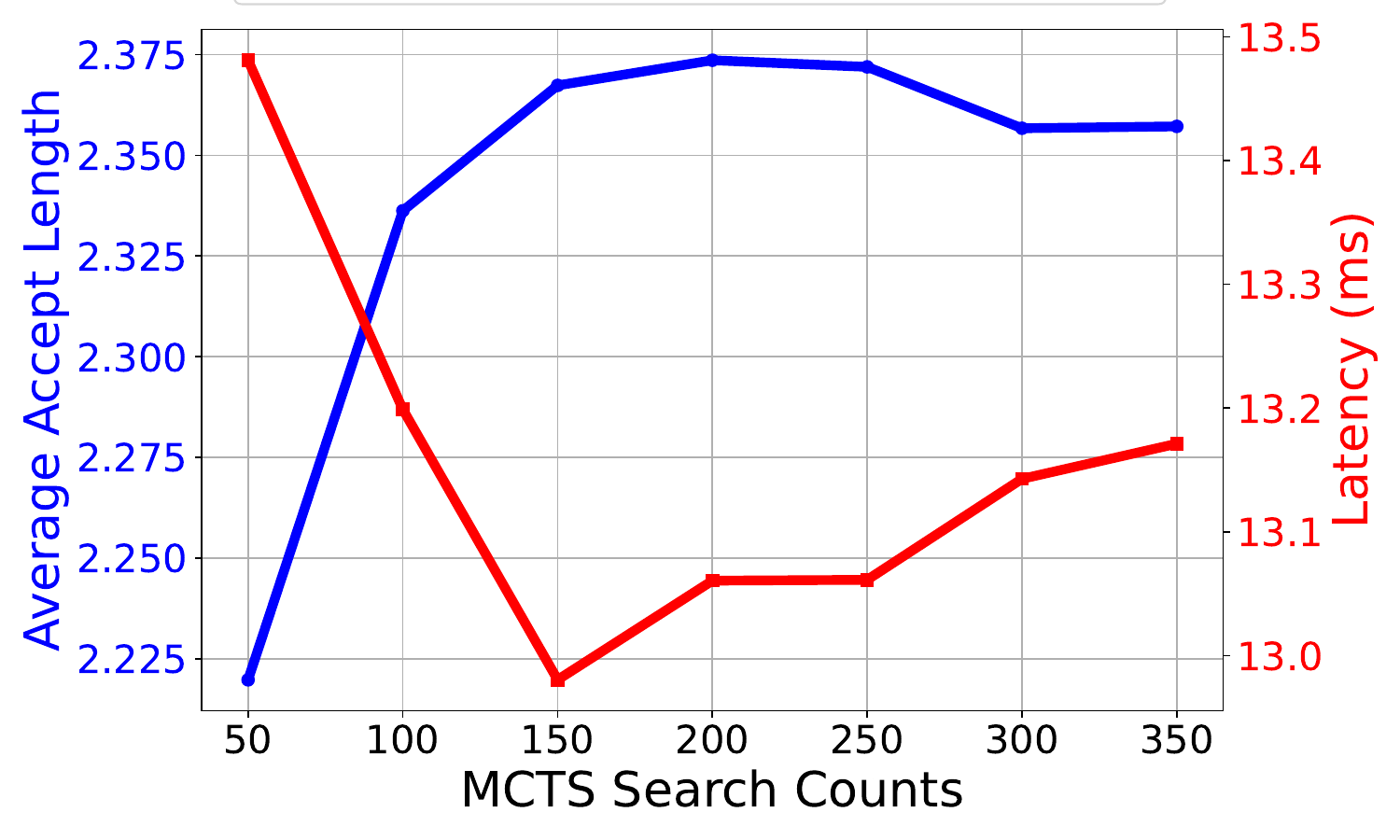}
        \caption{Vicuna-7B}
        \label{fig:sub11}
    \end{subfigure}
    \hfill  
    \begin{subfigure}[b]{0.49\columnwidth}
        \includegraphics[width=\textwidth]{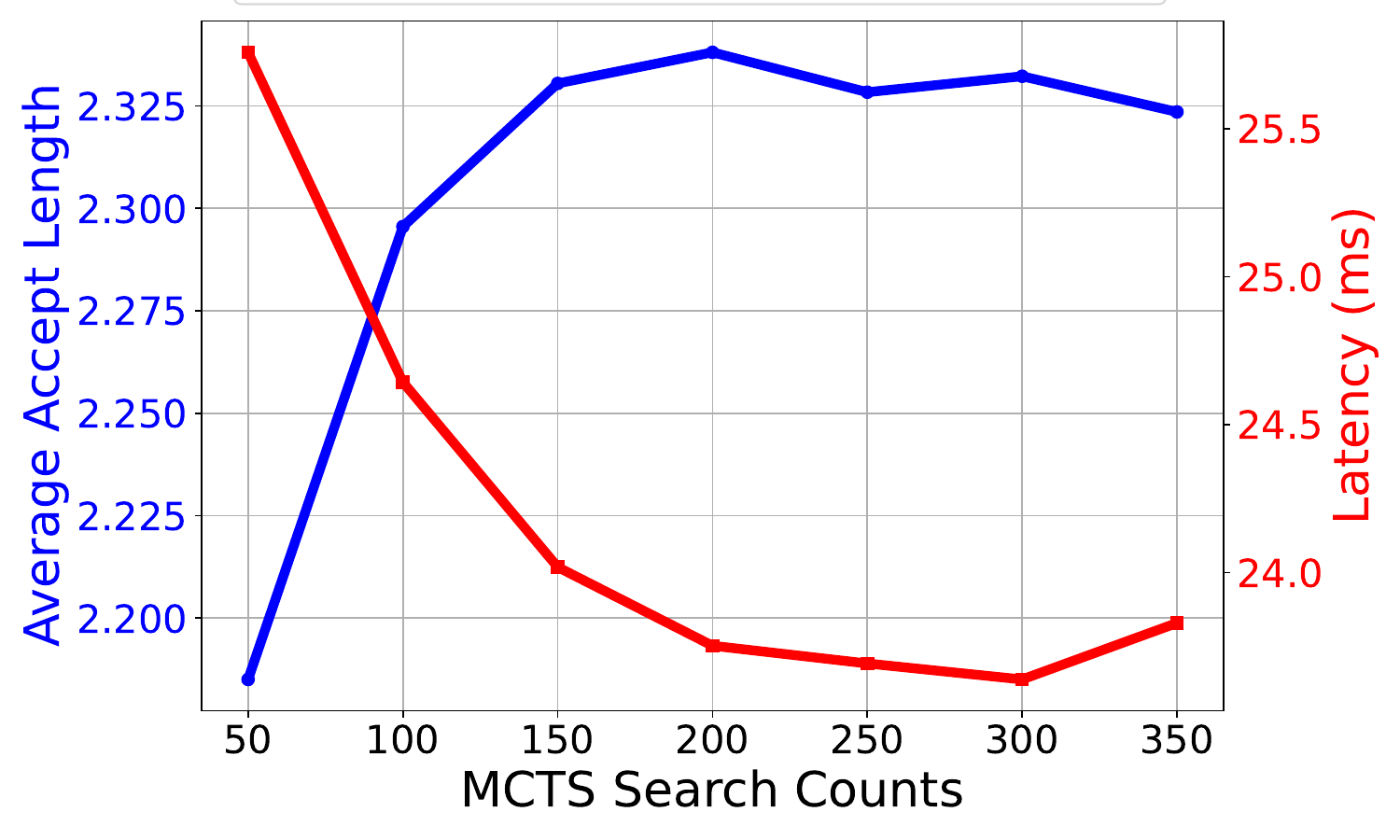}
        \caption{Vicuna-13B}
        \label{fig:sub22}
    \end{subfigure}
    \caption{Comparison of Search Counts and Performance on MT-Bench.}
    \label{fig:num}
\end{figure}

\textbf{Effect of \mcts.}
Figure~\ref{fig:num} presents the results for Vicuna-7B and Vicuna-13B models on the MT-Bench dataset, showing the impact of different \mcts search counts on performance. We find that for both models, increasing the number of searches improves performance, while the optimal number varies by model size. The average exact length and latency are plotted against the number of searches, illustrating the trade-off between performance and computational cost. 
As shown in the figure, there is a notable improvement in the average exact length as the number of searches increases, along with an increase in latency, indicating a balance between search depth and generation time.
We further compare greedy search with \mcts (full traversal is not feasible due to the huge search space) while keeping the number of 150 search iterations constant.
The results show that the average accept length of the greedy search is only 1.493, significantly lower than that obtained by MCTS, demonstrating the superiority of MCTS in efficiently managing the vast decision spaces.

\textbf{Effect of N-gram Model Choice.}
Our studies extensively evaluate the impact of different n-gram configurations on decoding performance. In tests conducted on the MT-Bench dataset using the Vicuna-7B model, bi-grams and 4-grams result in accept lengths of 1.80 and 1.82, respectively. These results are significantly lower compared to the 2.30 accept length achieved with tri-grams. 
Bi-gram models demonstrate limited capability in effectively utilizing contextual information, often leading to outputs that appear more random and less coherent. Conversely, 4-grams exhibit overly deterministic behavior, constraining the diversity of the generated text due to their restrictive nature in capturing extensive prior context. Tri-grams strike an optimal balance, providing enough contextual depth to enhance the coherence and relevance of outputs while still allowing for sufficient variability and diversity in text generation. This balance makes tri-grams particularly effective in large language model decoding, as they encapsulate the ideal compromise between randomness and contextual awareness.

%% file: sections/5-conclusion.tex
\section{Related Work}
A number of research efforts on decoding strategies for large language models have used draft models to improve decoding efficiency. Techniques such as Speculative Decoding~\citep{specd,spec1,spec2,spec3}, Madusa~\cite{medusa}, Eagle~\cite{eagle}, various other approaches requiring draft models~\citep{medusa2,sp1,sp2,sp3} fall into this category, utilizing models to generate drafts. Specifically, \textit{Speculative Decoding} uses an advanced sampling technique where the auxiliary model generates a set of potential token sequences, and the primary model selects the most sequences, resulting in a good balance between speed and accuracy. Although these methods primarily aim to enhance the accuracy of generated texts and significantly accelerate the response time during initial text generation, their adoption comes with drawbacks. The primary issue is the necessity for additional training specific to the draft models, which could be resource-intensive. Moreover, these techniques generally depend on GPU resources~\citep{gpu,gpu2,gpu1} for inference, potentially limiting their application in environments where such hardware is unavailable or when operating under strict resource constraints. 

A significant portion of recent advances has focused on improving efficiency without relying on draft models~\cite{look,he-etal-2024-rest}.
Two notable approaches in this realm are \textit{Lookahead decoding}~\cite{look} and \textit{Retrieval-Based Speculative Decoding} (REST)~\cite{he-etal-2024-rest}.
\textit{Lookahead decoding} is an approach that enhances the efficiency of the decoding process through the prediction of subsequent tokens via Jacobi Iteration~\cite{jacobi}. It employs a heuristic to estimate the future cost of a sequence without the need to explicitly create a draft. 
\textit{REST} introduces a retrieval-enhanced generation model that speculatively decodes sequences without the need for producing preliminary drafts. It instead searches and prioritizes possible continuations from an already established sequence database. 
However, these methods exhibit lower accuracy and greater resource use compared to our approach. They demand more memory and GPU processing power, posing challenges in resource-scarce settings.

\section{Conclusion}

\sysname improves the LLM decoding process by introducing adaptability and efficiency, significantly reducing latency and computational demands. This method achieves up to a 2.5X speedup in decoding and a 20\% improvement in acceptance rates, outperforming traditional techniques. Unlike existing approaches, \sysname dynamically adjusts the draft distribution using a tri-gram matrix and enhances draft quality through \mcts, eliminating the need for fine-tuning. The continuous feedback loop ensures ongoing improvements in draft generation. 
While \sysname demonstrates robust performance across various benchmarks, future work will focus on 
exploring its application in more diverse real-world scenarios. Additionally, addressing potential limitations in extremely large-scale deployments will be a priority.

%% file: sections/6-appendix.tex
\section*{\Large Appendix}
\appendix

\section{Advantages on Computation Efficiency}

\begin{table*}[t]
    \centering
    \caption{Comparison of Different Methods.}
    \begin{tabular}{lccc}
        \toprule
        \textbf{Method} & \textbf{Requires GPU} & \textbf{Computation} & \textbf{Memory Overhead} \\
        \midrule
        Lookahead & \textcolor{red}{\checkmark} & \textcolor{red}{$\uparrow$} & \textcolor{green}{$\times$} \\
        Eagle & \textcolor{red}{\checkmark} & \textcolor{red}{$\uparrow$} & \textcolor{green}{$\times$} \\
        Medusa & \textcolor{red}{\checkmark} & \textcolor{red}{$\uparrow$} & \textcolor{green}{$\times$} \\
        REST & \textcolor{green}{$\times$} & \textcolor{green}{$\downarrow$} & \textcolor{red}{$\uparrow$} \\
        Speculative Decoding & \textcolor{red}{\checkmark} & \textcolor{red}{$\uparrow$} & \textcolor{green}{$\times$} \\
        \sysname & \textcolor{green}{$\times$} & \textcolor{green}{$\downarrow$} & \textcolor{green}{Very Low} \\
        \bottomrule
    \end{tabular}
\end{table*}
Our technique provides substantial benefits when implemented on edge devices like laptops and smartphones, which often face limitations in GPU capabilities and memory. In contrast to traditional decoding methods that depend heavily on GPU power or large memory sizes, our strategy is crafted for high efficiency with low resource demands.

\textbf{Reduced GPU Requirements~} Our approach, which does not require fine-tuning and utilizes a lightweight probabilistic model, primarily operates on the CPU, eliminating the need for substantial GPU resources. This feature is especially advantageous for edge devices with limited GPU access. By minimizing GPU dependency, our technique can be applied more widely, enhancing LLM decoding across a broader array of devices.

\textbf{Low Memory Usage~} Our method avoids the need for bulky initial models or intricate neural network architectures, considerably lowering the memory usage typically needed for LLM decoding. This aspect is particularly suitable for devices with limited memory, such as budget laptops and mobile phones. The decrease in memory usage not only leads to quicker processing times but also reduces power consumption, which is vital for devices running on batteries. Compared to REST, which also requires a corpus, our method significantly reduces memory usage; for instance, both using the Stack dataset, our method requires only less than 1GB while REST needs 27GB.

Ultimately, our decoding method is exceptionally apt for practical use in edge systems, where there is often a scarcity of computational resources. It offers a viable and effective option for improving LLM decoding without sacrificing speed or precision, thus bringing sophisticated language processing to less powerful devices.

\section{Broader Impacts}

The advancements presented in this paper, specifically the accelerated LLM decoding via Monte Carlo Tree Search (\textit{\mcts}) and self-evolving speculation, have several broader impacts worth discussing. These impacts span multiple domains including technology, society, and ethics.

\subsection*{Technological Impact}
Our method significantly enhances the efficiency and speed of autoregressive LLM decoding. This improvement can benefit numerous applications that rely on real-time language processing, such as interactive chatbots, automated customer service, and real-time translation systems. By reducing the computational load and memory requirements, our approach also makes it feasible to deploy advanced LLMs on edge devices like smartphones and IoT devices, broadening their accessibility and usability.

\subsection*{Societal Impact}
The ability to perform faster and more efficient language model decoding can have a profound impact on society. For instance, it can improve the responsiveness and accuracy of assistive technologies for individuals with disabilities, such as voice-controlled assistants and text-to-speech systems. Additionally, educational tools that rely on real-time feedback and interactive learning can benefit from quicker and more reliable LLM responses, enhancing the learning experience for students.

\subsection*{Ethical Considerations}
While our advancements offer significant benefits, they also raise important ethical considerations. The increased efficiency of LLMs could lead to more widespread use of automated systems, which might replace human jobs in certain sectors. It is crucial to address the potential displacement of workers by fostering skills development and creating new job opportunities that leverage human-LLM collaboration.

Moreover, the deployment of more powerful LLMs on a wider scale necessitates robust measures to mitigate misuse. Enhanced LLM capabilities could be exploited for malicious purposes, such as generating misleading information or deepfake content. Therefore, it is essential to implement strong ethical guidelines and monitoring mechanisms to prevent abuse and ensure that the technology is used responsibly.

\subsection*{Environmental Impact}
Improving the efficiency of LLM decoding can also contribute to environmental sustainability. By reducing the computational resources required for LLM operations, our method decreases the energy consumption associated with running these models. This reduction is particularly important given the growing concerns about the environmental footprint of large-scale AI systems. Our approach aligns with the broader goal of developing greener AI technologies that minimize their impact on the planet.

In summary, the proposed method for accelerating LLM decoding has far-reaching implications across various domains. While it offers substantial benefits, it is essential to address the accompanying ethical, societal, and environmental challenges to ensure that the technology is developed and deployed in a responsible and beneficial manner.

\section{Limitations}
The \sysname methodology, while effective in accelerating LLM decoding, presents certain limitations primarily concerning accuracy and efficiency. Firstly, the retrieval-based approach integral to \sysname, although less resource-intensive, does not yet match the accuracy levels of traditional draft model methods. This discrepancy highlights a potential trade-off between resource efficiency and optimal accuracy in language model decoding. Secondly, the use of Monte Carlo Tree Search (\mcts) in our adaptive draft-verification process, though instrumental in improving decoding speed, still incurs additional computation. As a result, while \sysname achieves notable speed enhancements, the acceleration ratio remains suboptimal when compared to the theoretical maximum acceptance length. Future improvements will aim to refine these aspects to better balance speed and accuracy, potentially through more advanced optimization of the \mcts algorithm or alternative adaptive strategies.

\section{Ablation Study}
\subsection{Effect of Draft Search Score}
In this section, we present additional experiments to elucidate our choice and the implications of the hyperparameters $c_1$ and $c_2$ in the Draft Search Score formula.

Our analysis is grounded in Equation \ref{eq:e2}, which demonstrates that the exploration-exploitation balance is modulated by $c_1$ and $c_2$. Specifically, $c_1$ positively correlates with the exploration constant $E$, while $c_2$ inversely affects it.
\begin{figure}[htb]
    \centering
    \begin{subfigure}[b]{0.49\columnwidth}
        \includegraphics[width=\textwidth]{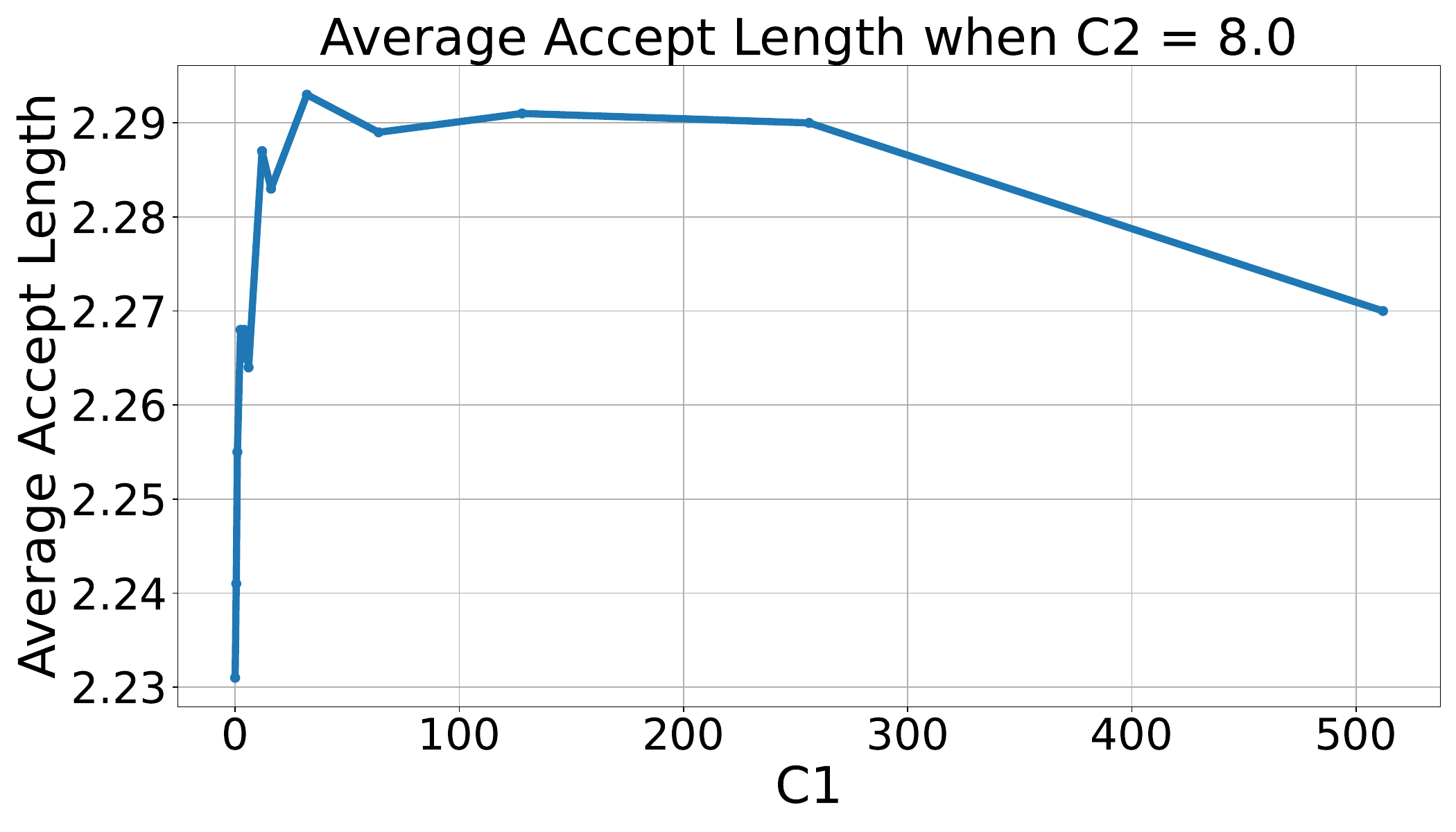}
        \caption{Impact of $c_1$ on Acceptance Length}
        \label{fig:c1}
    \end{subfigure}
    \hfill  
    \begin{subfigure}[b]{0.49\columnwidth}
        \includegraphics[width=\textwidth]{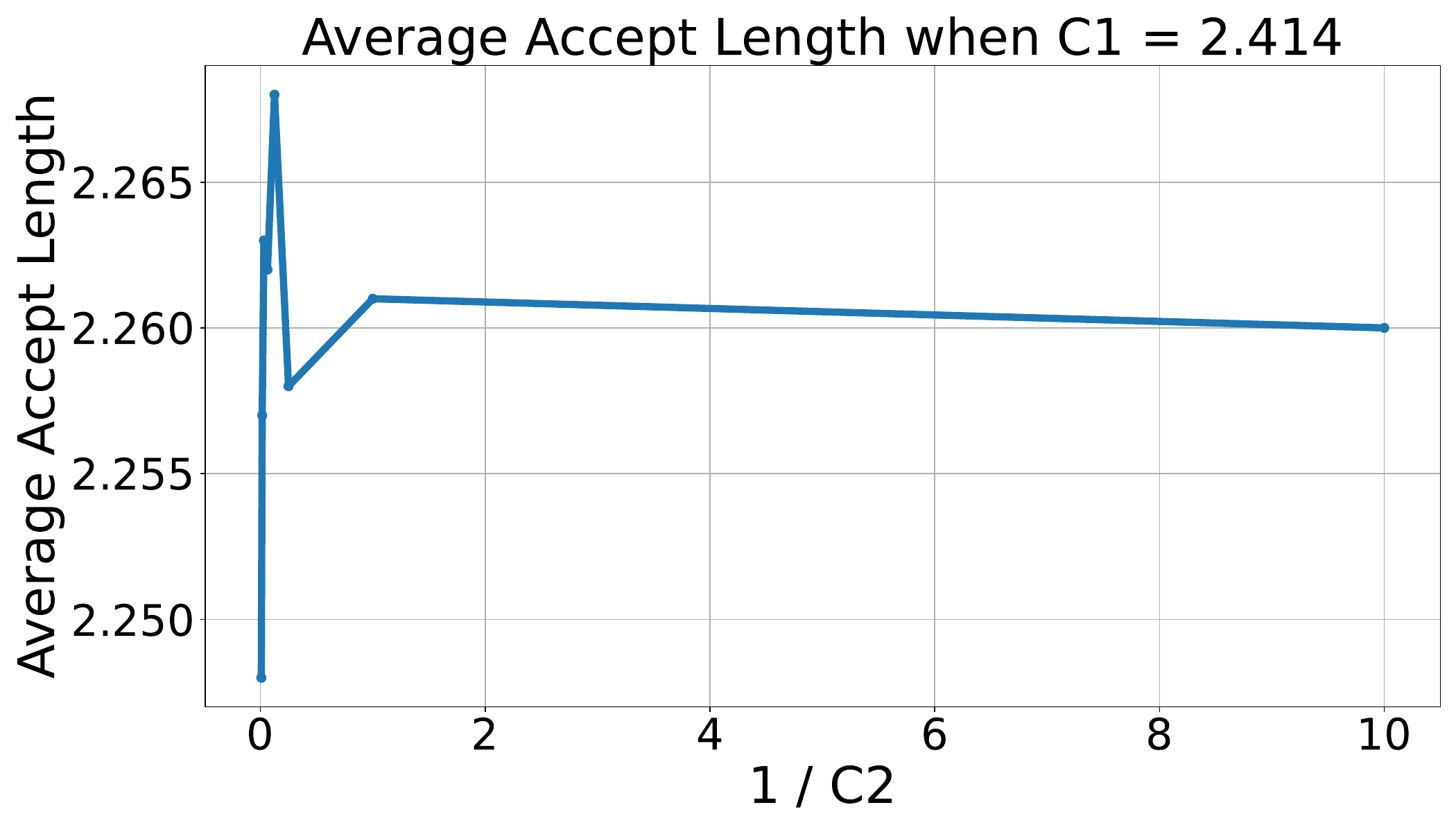}
        \caption{Impact of $c_2$ on Acceptance Length}
        \label{fig:c2}
    \end{subfigure}
    \caption{Relationship between Average Acceptance Length and Exploration Parameters.}
    \label{fig:c1c2}
\end{figure}

Figure \ref{fig:c1c2} illustrates the relationship between the average acceptance length and both $c_1$ and the reciprocal of $c_2$ ($1/c_2$). We observe an initial increase in performance with moderate levels of exploration, followed by a decline as the parameters extend to either extreme. This pattern suggests that both insufficient and excessive exploration can detrimentally affect model performance. This pattern indicates that both insufficient and excessive exploration levels impact the algorithm's performance, highlighting the importance of careful adjustment of $c_1$ and $c_2$.

\subsection{Summary}
In Experiment section, we conduct ablation studies to explore the impact of various configurations and optimizations in \sysname. 

The adaptive strategy shows significant benefits across different models, notably improving accept lengths by dynamically adjusting probability distributions based on tri-gram frequencies. 

Our examination of the effects of corpus size indicates that larger datasets contribute to improved accuracy, enhancing the coherence and quality of the output. However, even with smaller corpus sizes, our system demonstrates a notable increase in processing speed, underscoring its efficiency and capability to deliver accelerated performance without compromising on resource constraints.

The effectiveness of the Monte Carlo Tree Search (\mcts) is analyzed through varying search counts, demonstrating that increased iterations enhance performance while balancing computational costs. 

Additionally, comparisons between different n-gram models indicate that tri-grams achieve an optimal balance of randomness and contextual relevance, significantly outperforming other configurations in terms of decoding performance. 

These studies collectively affirm the critical design choices made in \sysname, validating its robustness and efficiency in language model decoding.

\section{Preliminary}


\textbf{Retrieval-Based Speculative Decoding:~} Decoding in large language models describes the procedure of text generation by sequentially predicting tokens. Given a context sequence $s = (x_1, ..., x_{t-1}, x_t)$, conventional autoregressive decoding techniques produce the subsequent token at position $t+1$ using conditional probability: 
\begin{equation} 
x_{t+1} \sim p(x | x_1, ..., x_t; \theta_{\text{large}}), 
\end{equation} 
where $p$ denotes the conditional probability distribution calculated by the LLM with parameters $\theta_{\text{large}}$. To reduce these computational burdens during inference, speculative decoding is proposed~\cite{specd}. It reduces the frequency of forward passes using $\theta_{\text{large}}$ by incorporating an auxiliary language model with fewer parameters $\theta_{\text{small}}$. 

The speculative decoding process is implemented iteratively as follows:
Using the smaller model $\theta_{\text{small}}$, a draft sequence of the next $m$ tokens is generated autoregressively:
\begin{equation}
\tilde{x}_{t+i} \sim p(x | s, \tilde{x}_{t+1}, ..., \tilde{x}_{t+i-1}; \theta_{\text{small}}),
\end{equation}
where $i = 1, ..., m$. Despite the sequential nature of this generation, the reduced model complexity of $\theta_{\text{small}}$ leads to a lower computational overhead compared to $\theta_{\text{large}}$.Retrieval-Based Speculative Decoding extends the basic speculative decoding framework by replacing the smaller language model with a retrieval system. This method uses:
\begin{equation}
\tilde{x}_{t+i} \sim p(x | s, \tilde{x}_{t+1}, ..., \tilde{x}_{t+i-1}; \theta_{\text{Corpus}}),
\end{equation}
where $\text{retrieve}(x_1, ..., x_t)$ fetches contextually relevant text segments from a pre-stored corpus, reducing reliance on frequent recalculations with $\theta_{\text{large}}$.

\textbf{Monte Carlo Tree Search:~}
Monte Carlo Tree Search (\mcts)~\cite{mcts} is a popular algorithmic in artificial intelligence, which explores potential future states from a current decision point by building a tree of possibilities, balancing exploration of new paths and exploitation of known beneficial paths. 

In the context of \mcts, each node in the tree represents a possible state, and these nodes are expanded based on the results of simulated plays. A key aspect of \mcts is how it selects nodes for further exploration, which is based on the number of visits to each node, denoted as \( N(s) \). Specifically, the selection process in \mcts can be interpreted as an approximate solution to a regular optimization problem\cite{mctsRPO}. The node visit count \( N(s) \) acts as a regularizer, guiding the algorithm towards a balance between exploring less visited, uncertain nodes and exploiting nodes that are known to yield higher rewards. The optimization problem can be formally expressed as
\begin{equation}
\max_{s \in \text{Children}(s')} \left( Q(s) + c \sqrt{\frac{\log N(s')}{N(s)}} \right),
\end{equation}
where \( s' \) is the current node, \( s \) represents a child node, \( Q(s) \) is the estimated value of node \( s \), \( N(s') \) is the number of visits to the parent node, and \( c \) is a constant that controls the trade-off between exploration and exploitation. This formulation highlights how \mcts inherently balances the dual objectives of accuracy (through \( Q(s) \)) and robustness (through the regularization term involving \( N(s) \)).

\textbf{PUCT Score:~}
The Probabilistic Upper Confidence Bound applied to Trees (PUCT) score is an adaptation of the Upper Confidence Bound (UCB) used in Monte Carlo Tree Search to optimize the exploration-exploitation trade-off in decision-making environments. The UCB formula is given by:
\begin{equation}\label{eq:ucb}
\text{UCB} = Q(s, a) + \sqrt{\frac{2 \log N}{N(s, a)}},
\end{equation}
where \(Q(s, a)\) represents the average reward of taking action \(a\) in state \(s\), \(N\) is the total number of trials, and \(N(s, a)\) is the number of trials where action \(a\) was taken from state \(s\).

To enhance the decision-making in tree-based strategies, the UCB was extended to Upper Confidence Bound applied to Trees (UCT), which adjusts the exploration parameter dynamically:
\begin{equation}\label{eq:uct}
\text{UCT} = Q(s, a) + C \cdot \sqrt{\frac{\log N}{N(s, a)}},
\end{equation}
where \(C\) is a variable exploration coefficient, adjusting dynamically based on the tree's exploration needs.

The PUCT score further incorporates probabilistic assessments into the UCT framework, allowing for decisions that consider the variance in outcome distributions:
\begin{equation}\label{eq:puct}
\text{PUCT} = Q(s, a) + c \cdot P(s, a) \cdot \sqrt{\frac{\log N(s)}{1 + N(s, a)}},
\end{equation}
where \(c\) is a constant scaling the exploration term, and \(P(s, a)\) represents the prior probability of selecting action \(a\) in state \(s\), enhancing the traditional exploration term by accounting for underlying probability distributions.

In our implementation, we use a variant of the original PUCT formula that allows for finer granularity in adjustments:
\begin{equation}\label{eq:e2}
E = C_1+\log\left(\frac{\sum_{b}N(s,b) + C_2 + 1}{C_2}\right),
\end{equation}
\begin{equation}\label{eq:mcts2}
 \text{PUCT}(s, a) = Q(s,a) + E \cdot P(s,a) \cdot \frac{\sqrt{\sum_{b}N(s,b)}}{1 + N(s,a)}.
\end{equation}
This adaptation allows our model to dynamically calibrate the exploration-exploitation balance more effectively, particularly in environments with large and complex decision spaces such as those involved in LLM decoding. The parameters \(c_1\) and \(c_2\) are introduced to further refine the balance between exploration and exploitation, where \(c_1\) scales the exploration bonus and \(c_2\) stabilizes the logarithmic term for numerical stability. This nuanced approach enhances the adaptability and efficiency of our \mcts framework, providing a substantial improvement over standard UCB and UCT methods.



\section{Detailed Experimental Settings}

\subsection{Configuration of \sysname}
For our experiments, we use the following hyperparameters to optimize the performance of \sysname: We set $t$, the threshold for the elimination of tri-gram probability, at 12 to focus only on the most prevalent tri-grams. The number of iterations for Monte Carlo Tree Search (\mcts), denoted $s$, is fixed at 150. The parameters in the \mcts PUCT Score function, $c_1$ and $c_2$, are 32.0 and 8.0, respectively, to effectively balance exploration and exploitation. The search depth and the length of each retrieved continuation candidate, represented by $l$, is 4; and the number of continuation candidates, $n$, is set at 24.

\subsection{Experimental Settings}
\begin{itemize}
    \item For all experiments listed in Table 1, \sysname uses UltraChat as the corpus, with the same configuration as mentioned in the previous section. All data presented in the table are the medians of three repeated runs.

    \item For Figure 4a, \sysname uses ShareGPT as the corpus with $c_1 = 2.414$.

    \item For Figure 4b, \sysname employs ShareGPT as the corpus.

    \item For Figure 5, \sysname utilizes ShareGPT as the corpus with $c_1 = 2.414$, and experiments are repeated seven times.

    \item For Table 2, \sysname is tested using two different corpora: ShareGPT (corpus size 121k) and UltraChat (corpus size 774k), with results reported as the median of three repetitions.

    \item For Figure 5, \sysname employs ShareGPT as the corpus, setting $c_1 = 2.414$.
\end{itemize}
\section{Configuration of Baselines}
\textbf{REST}
The REST baseline uses the default settings with the following specific configurations:
\begin{itemize}
    \item Number of threads: 6
    \item Draft choice: 64
    \item Datasets: UltraChat and The Stack
    \item Token spans: [16, 15, 14, 13, 12, 11, 10, 9, 8, 7, 6, 5, 4, 3, 2]
\end{itemize}

\textbf{REST Single Thread}
The REST Single Thread baseline uses the same settings as REST except the Number of threads:
\begin{itemize}
    \item Number of threads: 1
    \item Draft choice: 64
    \item Datasets: UltraChat and The Stack
    \item Token spans: [16, 15, 14, 13, 12, 11, 10, 9, 8, 7, 6, 5, 4, 3, 2]
\end{itemize}

\textbf{Lookahead}
The Lookahead baseline uses the default settings with the following specific configurations:
\begin{itemize}
    \item LEVEL: 5
    \item WIN: 15
    \item GUESS: 15
    \item FLASH: 0
\end{itemize}

\section{Implementation Details}
In this section, we describe the implementation details of our proposed Monte Carlo Tree Search (\mcts) algorithm for text generation and the dynamic adjustment of tri-gram probabilities.

\subsection{Monte Carlo Tree Search for Text Generation}
\begin{algorithm}[H]
\caption{Monte Carlo Tree Search for Text Generation}
\label{alg:mcts}
\textbf{Input}: $iterations$ — Number of iterations for \mcts, $self$ — The object containing the tree structure and parameters\\
\textbf{Output}: $sentences$ — Generated sentences ranked by their scores
\begin{algorithmic}[1]
\STATE $rng \gets \text{initialize random number generator}$
\FOR{$iter \gets 1$ \textbf{to} $iterations$}
    \STATE $node \gets self.root$
    \STATE $state \gets [node.word]$
    \STATE $end \gets \text{False}$
    \COMMENT{Selection}
    \WHILE{$\text{not } node.untried\_words \text{ is empty}$ \textbf{and} \\
       $node.children \text{ exist}$ \textbf{and} \\
       $\text{not } end$}

        \STATE $selected\_child \gets node.select\_child()$
        \STATE $state.append(selected\_child.word)$
        \STATE $node \gets selected\_child$
        \IF{$\text{length of } state == self.sentence\_length$}
            \STATE $end \gets \text{True}$
        \ENDIF
    \ENDWHILE
    \COMMENT{Expansion}
    \IF{$\text{not } node.untried\_words \text{ is empty and not } end$}
        \STATE $untried\_words \gets node.untried\_words$
        \STATE $p\_word \gets node.word.1$
        \FOR{$word \text{ in } untried\_words$}
            \STATE $tmp\_untried\_words \gets \text{get potential words from tri-gram matrix}$
            \STATE $child\_node \gets \text{new Node}(p\_word, word, tmp\_untried\_words)$
            \STATE $node.children.append(child\_node)$
        \ENDFOR
        \STATE $node \gets node.children.last()$
        \STATE $state.append(node.word)$
    \ENDIF
    \COMMENT{Simulation}
    \WHILE{$\text{length of } state < self.sentence\_length$}
        \STATE $last\_word \gets state.last()$
        \STATE $next\_words \gets \text{get from tri-gram matrix using } last\_word$
        \IF{$next\_words \text{ is not empty}$}
            \STATE $\text{select next word based on probabilities from }$
            \\ $next\_words$
            \STATE $state.append(\text{selected word})$
        \ELSE
            \STATE $\text{break}$
        \ENDIF
    \ENDWHILE
    \COMMENT{Backpropagation}
    \STATE $score \gets 1.0$
    \FOR{$i \gets 1$ \textbf{to} $\text{length of } state - 1$}
        \STATE $score \gets score + \text{get score from tri-gram matrix}$
    \ENDFOR
    \STATE $current\_node \gets node$
    \WHILE{$current\_node \text{ is not None}$}
        \STATE $current\_node.visits \gets current\_node.visits + 1$
        \STATE $current\_node.score \gets current\_node.score + score$
        \STATE $current\_node \gets current\_node.parent$
    \ENDWHILE
\ENDFOR
\COMMENT{Extract and sort top sentences}
\STATE $sentences \gets \text{extract sentences from root}$
\STATE $\text{sort } sentences \text{ by score}$
\STATE \textbf{return} $sentences$
\end{algorithmic}
\end{algorithm}

Algorithm ~\ref{alg:mcts} illustrates the Monte Carlo Tree Search (\mcts) procedure for text generation. The procedure, \textit{RunMCTS}, takes the number of iterations as input. The algorithm proceeds through the following phases:

\textbf{Selection:} Starting from the root node, the algorithm selects child nodes based on the selection criteria until a node with untried words or no children is reached, or the end of the sentence is detected.

\textbf{Expansion:} If the node has untried words and the end of the sentence is not reached, child nodes are created for each untried word using potential words from the tri-gram matrix.

\textbf{Simulation:} A sequence of words is generated starting from the current state until the sentence reaches the desired length or no more words can be selected from the tri-gram matrix.

\textbf{Backpropagation:} The score for the generated sequence is calculated, and this score is propagated back through the selected nodes, updating their visit counts and scores.

Finally, the top sentences are extracted from the root node and sorted by score.

\subsection{Dynamic Adjustment of tri-gram Probabilities}

Algorithm ~\ref{alg:3gram} details the procedure for dynamically adjusting tri-gram probabilities. The procedure, \textit{Adjust3Gram}, takes a sequence of tokens, the maximum length of n-grams to consider, an increment value, and the maximum probability.

For each tri-gram in the recent portion of the token sequence, the probability is increased by the increment value, up to the specified maximum probability. If the tri-gram is not already in the tri-gram matrix, it is added with the increment value as its initial probability.

\begin{algorithm}[H]
\caption{Dynamic Adjustment of tri-gram Probabilities}
\label{alg:3gram}
\textbf{Input}: $tokens$ — Sequence of tokens, $maxLength$ — Maximum length of tokens to consider\\
\textbf{Parameter}: $increment$ — Value by which the probability should be increased, $maxProb$ — Maximum allowable probability\\
\textbf{Output}: Updated tri-gram matrix with adjusted probabilities
\begin{algorithmic}[1]
\STATE $n \gets \text{length of } tokens$
\STATE $startIndex \gets \max(0, n - maxLength)$
\STATE $subTokens \gets tokens[startIndex:n]$
\FOR{$i \gets 2$ to $\text{length of } subTokens$}
    \STATE $triGram \gets (subTokens[i-2], subTokens[i-1], subTokens[i])$
    \IF{$triGram \text{ in tri-gram matrix}$}
        \STATE $currentProb \gets \text{tri-gram matrix}[triGram]$
        \STATE $newProb \gets \min(currentProb + increment, maxProb)$
        \STATE $\text{tri-gram matrix}[triGram] \gets newProb$
    \ELSE
        \STATE $\text{tri-gram matrix}[triGram] \gets increment$
    \ENDIF
\ENDFOR
\STATE \textbf{return} Updated tri-gram matrix
\end{algorithmic}
\end{algorithm}

\section{Detailed Explanation of Core Concepts and Interactions}
This section provides a detailed explanation of the relationships and interactions among the core components of our methodology: the Large Language Model (LLM), the LLM representative tri-gram matrix, Monte Carlo Tree Search (\mcts), and the final output decoding. These components are integral to our Adaptive Draft-Verification for Efficient LLM Decoding (\sysname) framework.

\subsection{LLM Representative Tri-gram Matrix}
The LLM representative tri-gram matrix is a crucial component of our \sysname framework. It serves as a condensed representation of the LLM's knowledge, capturing the conditional probabilities of token sequences based on their occurrence within the training corpus. This matrix is dynamically updated during the decoding process to reflect the changing context and to better predict subsequent tokens.

\subsection{Monte Carlo Tree Search (\mcts)}
\mcts is employed to explore potential draft outputs based on the probabilities indicated by the tri-gram matrix. It balances exploration of new, potentially high-reward token sequences with the exploitation of known, high-probability paths. This ensures a diverse yet accurate set of drafts that are likely to align closely with the true output distribution of the LLM.

\subsection{Interaction Between the Tri-gram Matrix and \mcts}
The tri-gram matrix provides the initial probability estimates that guide the \mcts in selecting the most promising token sequences during the draft generation phase. As \mcts explores different paths, it accumulates data about the success of various sequences, which in turn feedbacks into the tri-gram matrix. This feedback loop allows the tri-gram matrix to evolve and adapt, improving its accuracy and the efficiency of the draft verification process.

\subsection{Final Output Decoding}
Once the drafts have been generated and evaluated using \mcts, the final output decoding phase begins. This process involves comparing the drafts against the output distribution of the LLM using tree attention to select the sequence that best matches the expected outcomes. The selected draft then becomes the final output of the model for the given input context.

\subsection{Summary}
To aid in understanding, below is a conceptual diagram illustrating the interactions among these components. The following Figure ~\ref{fig:full_proc_all} shows the flow of data and decisions from the initial input through the tri-gram matrix, \mcts, and finally to the output.

\begin{figure*}[t]
    \centering
    \includegraphics[width=0.99\textwidth]{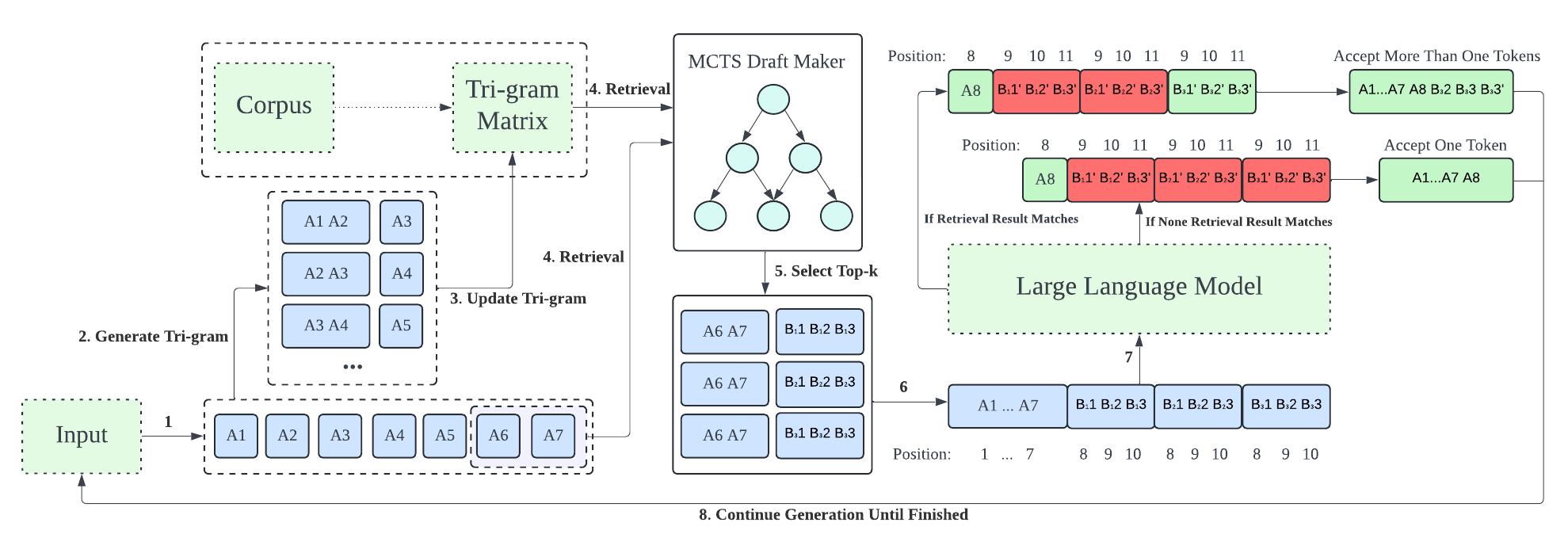}
    \caption{Data Flow and Interaction Among Core Algorithm Components. This figure details the sequential data processing steps within our \sysname. Initially, the input tokens are processed to compute their tri-grams, which are then utilized to update the tri-gram matrix. Subsequently, the updated tri-gram matrix, in conjunction with the last two tokens of the input, guides the Monte Carlo Tree Search (\mcts) in retrieving potential token sequences. The retrieval results are then ranked, and the top-k sequences are selected and appended to the original input. These extended sequences are fed into the Large Language Model (LLM) for further processing.}
    \label{fig:full_proc_all}
\end{figure*}

\begin{figure*}[t]
    \centering
    \includegraphics[width=0.99\textwidth]{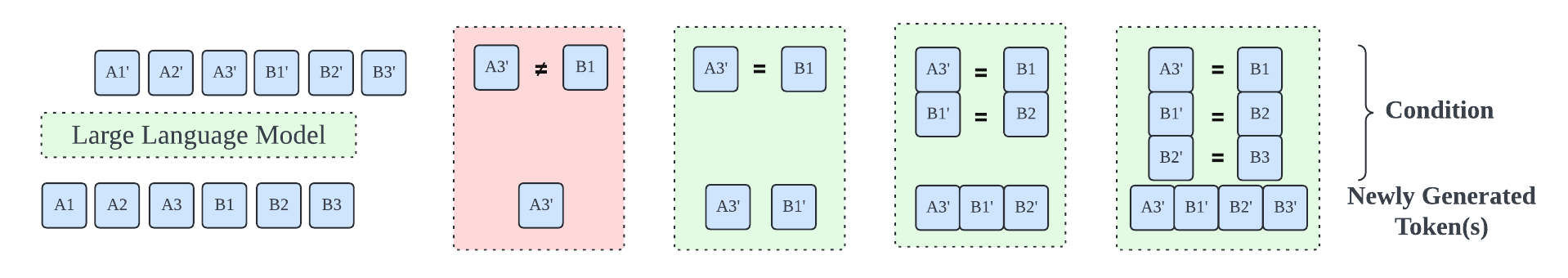}
    \caption{Validation of Retrieval Results in Large Language Model.}
    \label{fig:acc_check}
\end{figure*}

Additionally, Figure~\ref{fig:acc_check} complements this explanation, depicting the validation process used to assess the accuracy of retrieval results using tree attention.